\documentclass{style/new_tlp}
\usepackage{mathptmx}
\usepackage{listings}
\newcommand\bcmdtab{\noindent\bgroup\tabcolsep=0pt%
  \begin{tabular}{@{}p{10pc}@{}p{20pc}@{}}}
\newcommand\ecmdtab{\end{tabular}\egroup}

\let\proof\relax
 
\usepackage{tikz}
\usepackage{smartref}
\usetikzlibrary {positioning}
\usepackage{amsmath,amsthm}
\usepackage[notheorems,noamsthm,noalgnames,nodatasetnames,nocaptionpackage,noarraypackage]{style/pauli_new}
\usepackage{booktabs}

  \title[Theory and Practice of Logic Programming]
	{Hybrid ASP-based Approach to Pattern Mining}

  \author[S. Paramonov, D. Stepanova and P. Miettinen]
  {SERGEY PARAMONOV\\
   KU Leuven\\
    \email{sergey.paramonov@kuleuven.be}
    \and
    DARIA STEPANOVA and PAULI MIETTINEN\\
    Max Planck Institute for Informatics\\
    \email{\{dstepano,pmiettin\}@mpi-inf.mpg.de}}
\jdate{March 2017}
\pubyear{2017}
\pagerange{\pageref{firstpage}--\pageref{lastpage}}
\doi{S1471068401001193}
\usepackage{pifont}
\usepackage{multirow}
\usepackage{bbding}
\usepackage{amsmath}
\usepackage{amsthm}
\usepackage{amssymb}
\usepackage{wasysym}
\usepackage{xspace}
\usepackage{amstext}
\usepackage{mathtools}
\usepackage{paralist}
\usepackage{listings}
\usepackage{xspace}
\usepackage{subcaption}
\usepackage{fancyvrb}

\usepackage{graphicx}
\usepackage[utf8]{inputenc}
\usepackage[english]{babel}

\usepackage{tkz-tab}
\usepackage{caption}
\usepackage{latexsym}

\usetikzlibrary{arrows,matrix,positioning}

\newcommand{\leanparagraph}[1]{\smallskip\noindent\textbf{#1}. }
\newcommand{\rot}[1]{\rotatebox[origin=c]{-60}{#1}}
\newcommand{\na}{--}
\newcommand{\mi}[1]{\mathit{#1}}
\newcommand{\patternspace}{\ensuremath{\mathcal{L}}\xspace}
\newcommand{\subpattern}{\ensuremath{<^*}\xspace}
\newcommand{\cC}{\mathcal{C}}
\newcommand{\cI}{\mathcal{I}}
\newcommand{\cL}{\mathcal{L}}
\newtheorem{definition}{Definition}
\newtheorem{example}{Example}
\newcommand{\tuple}[1]{\langle#1\rangle}
\newcommand{\size}{\ensuremath{\textit{size}}\xspace}
\newcommand{\naf}[1]{\ensuremath{\mathit{not}~ #1}}
\providecommand{\url}[1]{\texttt{#1}}

\lstdefinestyle{model}{
     mathescape,
     columns=fullflexible,
     numbers=none,
     frame=single,
     escapechar=@
}

\newcommand{\commenttextasp}[1]{\textcolor{gray}{\textit{#1}}\xspace}
\lstdefinelanguage{ASPLang}
{
  morekeywords={
    import,
    if,
    while,
    for
  },
  sensitive=false, 
  morecomment=[l]{//}, 
  morecomment=[s]{/*}{*/}, 
  morestring=[b]" 
}

\newcommand{\qone}{\ensuremath{\textbf{Q}_\textbf{1}}\xspace}
\newcommand{\qtwo}{\ensuremath{\textbf{Q}_\textbf{2}}\xspace}
\newcommand{\qthree}{\ensuremath{\textbf{Q}_\textbf{3}}\xspace}
\newcommand{\qfour}{\ensuremath{\textbf{Q}_\textbf{4}}\xspace}
\newcommand{\qfive}{\ensuremath{\textbf{Q}_\textbf{5}}\xspace}
\newcommand{\qsix}{\ensuremath{\textbf{Q}_\textbf{6}}\xspace}

\newcommand{\comment}[1]{{\textcolor{blue}{*** #1 ***}}}

\newcommand{\revision}[1]{{#1}\xspace}
\newcommand{\doublerev}[1]{{#1}\xspace}

\begin{document}

\label{firstpage}

\maketitle

   \begin{abstract}
  Detecting small sets of relevant patterns from a given dataset is a central challenge in data mining. The relevance of a pattern is based on user-provided criteria; typically, all patterns that satisfy certain criteria are considered relevant. Rule-based languages like Answer Set Programming (ASP) seem well-suited for specifying such criteria in a form of constraints.  Although progress has been made, on the one hand, on solving individual mining problems and, on the other hand, developing generic mining systems, the existing methods either focus on scalability or on generality.  In this paper we make steps towards combining local (frequency, size, cost) and global (various condensed representations like maximal, closed, skyline) constraints in a generic and efficient way. We present a hybrid approach for itemset, sequence and graph mining which exploits dedicated highly optimized mining systems to detect frequent patterns and then filters the results using declarative ASP. To further demonstrate the generic nature of our hybrid framework we apply it to a problem of approximately tiling a database. Experiments on real-world datasets show the effectiveness of the proposed method and computational gains 
for itemset, sequence and graph mining, as well as approximate tiling. 

Under consideration in Theory and Practice of Logic Programming (TPLP).
\end{abstract}



\section{Introduction}\label{sec:intro}
\leanparagraph{Motivation} 
Availability of vast amounts of data from different domains has led to an increasing interest in the development of scalable and flexible methods for data analysis.
A key feature of flexible data analysis methods is their ability to incorporate users' background knowledge and different criteria of interest. They are often provided in the form of \emph{constraints} to the valid set of answers, the most common of which is the frequency threshold: a pattern is only considered interesting if it appears often enough. Mining all frequent (and otherwise interesting) patterns is a very general problem in data analysis, with applications in medical treatments, customer shopping sequences, Weblog click streams and text analysis, to name but a few examples.

Most data analysis methods consider only one (or few) types of constraints, limiting their applicability. Constraint Programming (CP) ~\cite{DBLP:conf/cpaior/NegrevergneG15,DBLP:journals/ai/GunsDNTR17} has been proposed as a general approach for (sequential) mining of frequent patterns~\cite{DBLP:books/mit/fayyadPSU96/AgrawalMSTV96}, and Answer Set Programming (ASP) \revision{\cite{GL1988,eiter} has been proven to be well-suited for defining the constraints conveniently (see e.g., \citeN{DBLP:conf/lpnmr/Jarvisalo11}, \citeN{DBLP:conf/ijcai/GebserGQ0S16} and \citeN{DBLP:journals/corr/GuyetMQ14} for existing approaches on ASP-based frequent pattern mining) thanks to its expressive and intuitive modelling language and the availability of optimized ASP solvers such as Clasp \cite{clasp} and WASP \cite{wasp}.}

In general, all constraints can be classified into \emph{local constraints}, that can be validated by the pattern candidate alone, and \emph{global constraints}, that can only be validated via an exhaustive comparison of the pattern candidate against all other candidates. Combining local and global constraints in a generic way is an important and challenging problem, which has been widely acknowledged in the constraint-based mining community.  \revision{The existing methods have focused either on scalability of solving individual mining problems or on generality, but rarely address both of these aspects.}  This naturally limits the practical applicability of the existing approaches.

\leanparagraph{State of the art and its limitations} Purely declarative ASP encodings for frequent and maximal itemset mining were proposed by \citeN{DBLP:conf/lpnmr/Jarvisalo11}. In this approach, every item's inclusion into the candidate itemset is guessed at first, and the guessed candidate pattern is checked against frequency and maximality constraints. While natural, this encoding is not  truly generic, as adding extra local constraints requires significant changes in it. Indeed, for a database where all available items form a frequent (and hence maximal) itemset, the maximal ASP encoding has a single model. The latter is, however, eliminated once restriction on the length of allowed itemsets is added to the program. This is undesired, as being maximal
is not a  property of an itemset on its own, but rather 
in the context of a collection of other itemsets \cite{DBLP:journals/kais/BonchiL06}. Thus, ideally one would be willing to first apply all local constraints and only afterwards construct a condensed representation of them, which is not possible in the approach of  \citeN{DBLP:conf/lpnmr/Jarvisalo11}. 

This shortcoming has been addressed in the recent work on ASP-based sequential pattern mining \cite{DBLP:conf/ijcai/GebserGQ0S16}, which exploits ASP preference-handling capacities to extract patterns of interest and supports the combination of local and global constraints. However, both \citeN{DBLP:conf/ijcai/GebserGQ0S16} and \citeN{DBLP:conf/lpnmr/Jarvisalo11} present 
purely declarative encodings, which suffer from scalability issues caused by the exhaustive exploration of the huge search space of candidate patterns \revision{(existing solvers cannot yet take the full advantage of the stucture of the problem \cite{KR_Graphs}, as specialized algorithms do to scale up and work with large datasets)}. The subsequence check amounts to testing whether an embedding exists (matching of the individual symbols) between sequences. In sequence mining, a pattern of size $m$ can be embedded into a sequence of size $n$ in $O(n^m)$ different ways, therefore, clearly a direct pattern enumeration is unfeasible in practice. 

While a number of individual methods tackling selective constraint-based mining tasks exist (see Table~\ref{tab:comparison} for comparison) there is no uniform ASP-based framework that is capable of effectively combining constraints both on the global and local level and is suitable for itemsets, sequences and graphs alike.

\leanparagraph{Contributions} The goal of our work is to make steps towards building a generic framework that supports  mining of condensed 
patterns, which (1) effectively combines dedicated algorithms and declarative means for pattern mining and (2) is easily extendable to incorporation of various constraints. 
\revision{We propose a two-step approach.  In the first step, optimized algorithms are applied to discover a set of frequent patterns, and in the second step, the patterns are post-processed using declarative means. The key advantage of our approach stems from the fact that it preserves the generality of purely declarative methods with respect to the frequent pattern mining problems of the specified types, while providing an efficient system to develop prototypes which can run on real-world datasets where typically only specialized algorithms are deployed. This is especially beneficial in the setting where a user considers a new variation of a pattern mining problem and needs to prototype a system to run on the standard real-world pattern mining datasets. Typically, even standard pattern mining datasets are too large for purely declarative systems, and researchers have to experiment with the smallest datasets available (see, for example, experimental sections of \citeN{DBLP:conf/ijcai/GebserGQ0S16}, \citeN{dp2013}, \citeN{DBLP:conf/cpaior/NegrevergneG15}, \citeN{k_pattern_mining_under_constraints} and \citeN{query_mining_ilp}); contrary to this, developers of specialized algorithms, practically, have to rewrite the algorithms almost completely to model and solve new variations of a problem (for example, see separate algorithms and papers for gSpan \cite{gspan} and cloSpan \cite{clospan}). Our approach provides a middle ground between them, on the one hand, a researcher can model a new problem variation without changing the whole model and, on the other hand, she can experiment with the real-world datasets, which indeed makes the declarative approach more practical and appealing for applications.}

The salient contributions of our work can be summarized as follows:

\begin{itemize}
  \item  We present a general extensible pattern mining framework for mining patterns of different types using ASP.
  \item In addition to the classical pattern mining problems, we demonstrate the generic nature of our framework by applying it to a problem of approximately tiling a database. 
  \item  \revision{We introduce a feature comparison between different ASP mining models and dominance programming (a generic itemset mining language and solver).} 
  \item  We demonstrate the feasibility of our approach with an experimental evaluation across multiple itemset, sequence and graph datasets using state-of-the-art ASP solvers.
\end{itemize}


 \leanparagraph{Structure} 
 After providing necessary background in Section~\ref{sec:prelim} we introduce our approach in Section~\ref{sec:problem}, and discuss approximate pattern mining in Section~\ref{sec:tiling}. Experimental results are described in Section~\ref{sec:eval}, while related work and final remarks are provided in Section~\ref{sec:relwork} and Section~\ref{sec:conc} respectively.

\begin{table}[t]
  \centering
  \setlength\tabcolsep{3.0pt}
  \begin{tabular}{@{}lccccc@{}}
    \toprule
    \textbf{Datatype}                & \textbf{Task}                  & \rot{\citeNP{DBLP:conf/lpnmr/Jarvisalo11}} & \rot{\citeNP{DBLP:conf/ijcai/GebserGQ0S16}} & \rot{\citeNP{dp2013}} &  \rot{\textbf{Our work}} \\  \midrule 
                                                                                                                                                                      
  \multirow{3}{*}{\textbf{Itemset}}  & frequent pattern mining        &  \checkmark      &  \na        & \checkmark       & \checkmark   \\ 
                                     & condensed (closed, max, etc)   & $\checkmark^{*}$ &  \na        & \checkmark       & \checkmark   \\ 
                                     & condensed under constraints    &  \na             &  \na        & \checkmark       & \checkmark   \\\midrule 
  \multirow{3}{*}{\textbf{Sequence}} & frequent pattern mining        &  \na             & \checkmark  & \na              & \checkmark   \\ 
                                     & condensed (closed, max, etc)   &  \na             & \checkmark  & \na              & \checkmark   \\ 
                                     & condensed under constraints    &  \na             & \checkmark  & \na              & \checkmark       \\\midrule 
 \multirow{3}{*}{\textbf{Graphs}}  & frequent pattern mining        &  \na     &  \na        &  \na      & \checkmark  \\ 
                                     & condensed (closed, max, etc)   & \na &  \na        & \na       & \checkmark   \\ 
                                     & condensed under constraints    &  \na             &  \na        & \na       & \checkmark   \\ \bottomrule 
                                                                                                                                           
  \end{tabular} 
 \caption{Feature comparison between various ASP mining models and dominance programming (``\na'' : ``not designed for this datatype'', $\checkmark^*$ : only maximal is supported)}
  \label{tab:comparison}
\end{table}


\section{Preliminaries}\label{sec:prelim}
In this section we briefly recap the necessary background both from the fields of pattern mining and Answer Set Programming (ASP). 

Let $\mi{D}$ be a dataset, $\patternspace$ a language for expressing pattern properties or defining subgroups of the
data, and $\mi{q}$ a selection predicate. 
The task of pattern mining is to find $\mi{Th}(\patternspace,D, \mi{q})= \{\phi \in \patternspace \mid \mi{q}(D, \phi) \text{ is true}\}$, that is, to find all patterns $\phi\in\patternspace$ that are selected by $q$  (see, e.g, the seminal work of \citeN{DBLP:journals/datamine/MannilaT97}). 

Pattern mining has been mainly studied for 
itemsets, sequences, graphs and tilings. These settings are determined by the language of $\patternspace$. In this work we discuss all of these pattern types.

\subsection{Patterns} 
\subsubsection{Itemsets}
 \emph{Itemsets} represent the most simple setting of frequent pattern mining. Let $\cI$
be a set of items $\{o_1,o_2,\dotsc,o_n\}$. A nonempty subset of $\cI$ is called an \emph{itemset}. A
\emph{transaction dataset} $D$ is a collection of itemsets, $D = \{t_1 ,\dotsc, t_m \}$, where $t_i\subseteq \cI$.
For any itemset $\alpha$, we denote the set of transactions that contain $\alpha$ as $D_{\alpha}= \{i\mid\alpha \subseteq t_i, t_i \in D\}$; we refer to $D_{\alpha}$ as the \emph{cover} of an itemset $\alpha$ and to $\abs{D_{\alpha}}$ as the \emph{support (frequency)} of $\alpha$ in $D$, written $\mi{sup}(\alpha)$. The \emph{relative frequency} of $\mi{\alpha}$ in $D$ refers to the ratio between $\mi{sup}(\alpha)$ and $\abs{D}$. 
The \emph{cardinality} (or \emph{size}) $\abs{\alpha}$ of an itemset $\alpha$ is 
the number of items contained in it. 

\begin{definition}[Frequent Itemset]\label{def:frit} 
 Given a transaction dataset $D$ and a frequency threshold $\sigma \geq 0$, an itemset $\alpha$ is \emph{frequent} in $D$ if  $\mi{sup}(\alpha)\geq \sigma$.\footnote{In \emph{frequent pattern mining}, often, a \emph{relative threshold}, i.e., $\sigma/\abs{D}$ is specified by the user.}
\end{definition}

We illustrate the introduced notions by the following example.

\begin{example}\label{ex:it}
Consider a transaction dataset $D$ from Table~\ref{tab:it}.
We have 
$\cI=\{a,b,c,d,e\}$ and $\abs{D}=3$. For $\sigma=2$, the following itemsets are frequent: $\alpha_1{=}\{a\}$, $\alpha_2{=}\{b\}$, $\alpha_3{=}\{e\}$, $\alpha_4{=}\{a,e\}$ and $\alpha_5{=}\{b,e\}$. Moreover, it holds that $D_{\alpha_4}=\{1,3\}, D_{\alpha_5}=\{1,2\}$, and the coverage for the rest of the itemsets can be analogously found. \qed
\end{example}

\begin{table}[t]
    \begin{minipage}{.5\linewidth}
      \centering
\begin{tabular}{|l|l|l|l|l|l|}
\hline
$\mi{ID}$& $\mi{a}$ &$\mi{b}$&$\mi{c}$&$\mi{d}$&$\mi{e}$\\ \hline
$\mi{1}$ & $\checkmark$ &$\checkmark$ &&$\checkmark$&$\checkmark$\\ \hline
$\mi{2}$ &  &$\checkmark$ &$\checkmark$&&$\checkmark$\\ \hline
$\mi{3}$ &$\checkmark$ &&&&$\checkmark$ \\ \hline
\end{tabular}
      \caption{Transaction database}
\label{tab:it}
    \end{minipage}%
    \begin{minipage}{.5\linewidth}
      \centering
\begin{tabular}{|l|l|}
\hline
$\mi{ID}$& Sequence\\ \hline
$\mi{1}$ & $\tuple{\mi{a\,b\,c\,d\,a\,e\,b}}$ \\ \hline
$\mi{2}$ & $\tuple{\mi{b\,c\,e\,b}}$\\ \hline
$\mi{3}$ & $\tuple{\mi{a\,a\,e}}$ \\ \hline
\end{tabular}
      \caption{Sequence database}
\label{tab:seq}

    \end{minipage} 
\label{tab:ex}
\end{table}

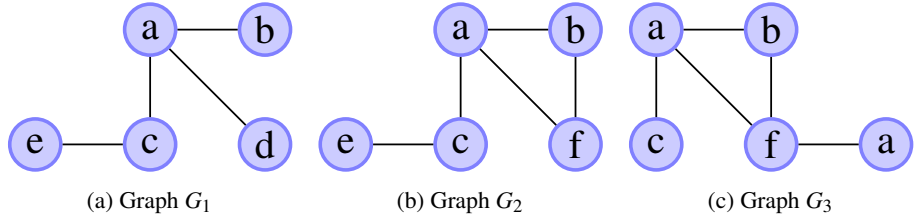
\begin{figure}
    \begin{subfigure}[b]{0.3\textwidth}
        \centering
        \resizebox{\linewidth}{!}{
       \begin{tikzpicture}[scale=0.9,
        circ/.style={circle,draw=blue!50,fill=blue!20,thick,
          inner sep=0pt,minimum size=4mm},
        transition/.style={rectangle,draw=black!50,fill=black!20,thick,
          inner sep=0pt,minimum size=4mm}]
      \node (1) at ( 0,2) [circ] {a};
      \node (2) at ( 1,2) [circ] {b};
      \node (3) at ( 0,1) [circ] {c};
      \node (4) at ( 1,1) [circ] {d};
      \node (5) at ( -1,1) [circ] {e};
      \draw (1) edge[-] (2);
      \draw (1) edge[-] (3);
      \draw (1) edge[-] (4);
      \draw (3) edge[-] (5);
\end{tikzpicture}

        }
        \caption{Graph $G_1$}
        \label{fig:g1}
    \end{subfigure}
    \begin{subfigure}[b]{0.3\textwidth}
    \centering
        \resizebox{\linewidth}{!}{
           
\begin{tikzpicture}[scale=0.9,
        circ/.style={circle,draw=blue!50,fill=blue!20,thick,
          inner sep=0pt,minimum size=4mm},
        transition/.style={rectangle,draw=black!50,fill=black!20,thick,
          inner sep=0pt,minimum size=4mm}]
      \node (1) at ( 0,2) [circ] {a};
      \node (2) at ( 1,2) [circ] {b};
      \node (3) at ( 0,1) [circ] {c};
      \node (4) at ( 1,1) [circ] {f};
      \node (5) at ( -1,1) [circ] {e};
      \draw (1) edge[-] (2);
      \draw (1) edge[-] (3);
      \draw (1) edge[-] (4);
      \draw (3) edge[-] (5);
      \draw (2) edge[-] (4);
\end{tikzpicture}

        }
        \caption{Graph $G_2$}   
        \label{fig:g2}
    \end{subfigure}
    \begin{subfigure}[b]{0.3\textwidth}
        \centering
        \resizebox{\linewidth}{!}{
  
\begin{tikzpicture}[scale=0.9,
        circ/.style={circle,draw=blue!50,fill=blue!20,thick,
          inner sep=0pt,minimum size=4mm},
        transition/.style={rectangle,draw=black!50,fill=black!20,thick,
          inner sep=0pt,minimum size=4mm}]
      \node (1) at ( 0,2) [circ] {a};
      \node (2) at ( 1,2) [circ] {b};
      \node (3) at ( 0,1) [circ] {c};
      \node (4) at ( 1,1) [circ] {f};
      \node (5) at ( 2,1) [circ] {a};
      \draw (1) edge[-] (2);
      \draw (1) edge[-] (3);
      \draw (1) edge[-] (4);
      \draw (4) edge[-] (5);
      \draw (2) edge[-] (4);
\end{tikzpicture}

        }
        \caption{Graph $G_3$}
        \label{fig:g3}
    \end{subfigure}
\bigskip

\caption{Graph examples} 
\label{fig:graphs}
\end{figure}

\subsubsection{Sequences} A \emph{sequence} is an ordered set of items $\tuple{s_1,\dotsc,s_n}$.
The setting of \emph{sequence mining} includes two related yet different cases: frequent substrings 
and frequent subsequences. In this work we focus on the latter.

\begin{definition}[Embedding in a Sequence]\label{def:embseq}
Let $S=\tuple{s_1,\dotsc,s_m}$ and $S'=\tuple{s_1',\dotsc,s_n'}$ be  two  sequences  of  size $m$ and $n$ respectively  with $m \leq n$.  The  tuple  of  integers $e=(e_1,\dotsc,e_m)$ is an \emph{embedding} of $S$ in $S'$ (denoted $S\sqsubseteq_e S')$ if and only if $e_1< \dotsc < e_m$ and for any $i \in \{1,\ldots,m\}$ it holds that $s_i=s'_{e_i}$.
\end{definition}

\begin{example}\label{ex:seq}
	For \revision{the} dataset in Table ~\ref{tab:seq} we have that $\tuple{b\,c\,e\,b} \sqsubseteq_{e_1} \tuple{\mi{a\,b\,c\,d\,a\,e\,b}}$ for $\mi{e_1}=(2,3,6,7)$ and analogously, $\tuple{\mi{a\,a\,e}}\sqsubseteq_{e_2} \tuple{\mi{a\,b\,c\,d\,a\,e\,b}}$ with $e_2=(1,5,6)$. \qed
\end{example}

We are now ready to define an inclusion relation for sequences. 

\begin{definition}[Sequence Inclusion]\label{def:seqinc}
Given two sequences $S = \tuple{s_1,\dotsc, s_m}$ and $S'=\tuple{s_1',\dotsc, s_n'}$, of 
size $m$ and $n$, respectively, with $m \leq n$, we say that $S$ is \emph{included} in $S'$ or $S$ is a \emph{subsequence} of $S'$ denoted by $S \sqsubseteq S'$ iff an embedding $e$ of $S$ in $S'$ exists, i.e.
\begin{equation}
S \sqsubseteq S' \leftrightarrow \exists e_1<\dotsc<e_m \text{ and }\forall i \in 1\dotsc m: s_i=s_{e_i}'.
\end{equation}
\end{definition}

\begin{example}
  In Example~\ref{ex:seq} we have  $\tuple{b\,c\,e\,b} \sqsubseteq \tuple{\mi{a\,b\,c\,d\,a\,e\,b}}$ but  $\tuple{\mi{a\,a\,e}} \not \sqsubseteq \tuple{\mi{b\, c\, e\, b}}$. \qed
\end{example}

For a given sequence $S$ and a sequential dataset $D=\{S_1,\dotsc,S_n\}$ we denote by $D_S$ the subset of $D$ s.t. $S \sqsubseteq S'$ for all $S' \in D_S$. The support of $S$ is $\mi{sup}(S)=\abs{D_S}$. Frequent sequences are defined analogously to frequent itemsets. 

\begin{definition}[Frequent Sequence]\label{def:frseq} Given a sequential dataset $D=\{S_1,\dotsc, S_n\}$ and a frequency threshold $\sigma \geq 0$, a sequence $S$ is \emph{frequent} in $D$ if $\mi{sup}(S)\geq \sigma$. 
\end{definition}

\begin{example}
	For \revision{the} dataset in Table~\ref{tab:seq} and $\sigma=2$, it holds that $\tuple{\mi{b\,c\,e\,b}}$ and $\tuple{a\,a\,e}$ are frequent, while $\tuple{\mi{b\,d\,b}}$ is not. \qed
\end{example}

Note that $\sqsubseteq$ and $\subseteq$ are incomparable relations. Indeed, consider two sequences $s_1=\tuple{a\,b}$ and $s_2=\tuple{b\,a\,a}$. While $s_1\subset s_2$, we clearly have that $s_1 \not \sqsubset s_2$.

\subsubsection{Graphs} A \emph{graph} $G$ is a triple $\tuple{V, E, l}$ where $V$ is a set of vertices, $E$ is a set of edges and $l$ is a labeling function that maps each edge and each vertex to a label. 

In this work we consider undirected graphs. Moreover, we primarily focus on two settings: a restricted one, where unique labels are ensured and a general one, where labels of the graph are not necessarily unique.
\revision{\paragraph{Uniquely Labelled Graphs.} In this restricted setting, we assume that each node has a unique label within the graph, and no labels on the edges are provided.    
This restriction makes the sub-pattern check computationally easier and allows certain reductions to other pattern mining problems as indicated below.}
\begin{example}
    The graphs in Figure~\ref{fig:g1} and \ref{fig:g2} are unique-labeled ones, while the graph from Figure~\ref{fig:g3} is clearly not. \qed
\end{example} \vspace{-10pt}
\revision{\paragraph{General Case of Graph Mining.} 
In the general case we consider 
graphs, whose nodes have labels, that are not necessarily unique. 
Here, a pattern is an arbitrary graph with labeled nodes and edges. This general case 
is 
computationally more demanding 
than the special case of uniquely labelled graphs.}

\revision{In the further exposition of the results, the general case is assumed by default, unless explicitly stated otherwise.}
\begin{definition}[Graph Isomorphism]\label{def:isom}
    Given two graphs $G=\tuple{V,E,l}$ and $G'=\tuple{V',E',l'}$, we say that $G$ is \emph{isomorphic} to $G'$ iff there is a \revision{bijective function} $f$ such that 
    \revision{
        \begin{itemize}
            \item $v \in V$ iff $f(v)\in V'$ and for all $v \in V$ it holds that $l(v)=l'(f(v))$
            \item $e \in E$ iff $f(e)\in E'$ and for all $e \in E$ it holds that $l(e)=l'(f(e))$.
        \end{itemize}
    }
\end{definition}
\revision{The graph isomorphism problem is claimed to be solvable in quasipolynomial time \cite{graph_isomorphism}.}
\begin{definition}[Graph Inclusion]\label{def:subgraph}
    Given two graphs $G = \tuple{V,E}$ and $H=\tuple{U, F}$, such that $|V|\leq |F|$, we say that $G$ is an \emph{(isomorphic) subgraph} of $H$  denoted by $G \sqsubseteq H$ iff there exists a subgraph $H' = \tuple{U'\subseteq U, F'\subseteq F}$ such that $G$ is isomorphic to $H'$.
\end{definition}

We now illustrate the introduce notions by the following example.

\begin{figure}
    \begin{subfigure}[b]{0.3\textwidth}
        \centering
        \resizebox{\linewidth}{!}{
            \begin{tikzpicture}[scale=0.9,
                circ/.style={circle,draw=blue!50,fill=blue!20,thick,
                inner sep=0pt,minimum size=4mm},
                transition/.style={rectangle,draw=black!50,fill=black!20,thick,
                inner sep=0pt,minimum size=4mm}]
                \node (1) at ( 0,2) [circ] {a};
                \node (2) at ( 1,2) [circ] {b};
                \node (3) at ( 0,1) [circ] {c};
                \node (5) at ( -1,1) [circ] {e};
                \draw (1) edge[-] (2);
                \draw (1) edge[-] (3);
                \draw (3) edge[-] (5);
            \end{tikzpicture}     
        }
        \caption{Graph $G'_1$}
        \label{fig:sg1}
    \end{subfigure}
    \begin{subfigure}[b]{0.3\textwidth}
        \centering
        \resizebox{\linewidth}{!}{
            \begin{tikzpicture}[scale=0.9,
                circ/.style={circle,draw=blue!50,fill=blue!20,thick,
                inner sep=0pt,minimum size=4mm},
                transition/.style={rectangle,draw=black!50,fill=black!20,thick,
                inner sep=0pt,minimum size=4mm}]
                \node (1) at ( 0,2) [circ] {a};
                \node (2) at ( 1,2) [circ] {b};
                \node (4) at ( 1,1) [circ] {f};
                \node at (-0.6,1) {};
                \node at (1.6,1) {};
                \draw (1) edge[-] (2);
                \draw (1) edge[-] (4);
                \draw (2) edge[-] (4);
            \end{tikzpicture}  
        }
        \caption{Graph $G'_2$}
        \label{fig:sg2}
    \end{subfigure}
    \caption{Isomorphic subgraphs for graphs in Figure~\ref{fig:graphs}} 
    \label{fig:subgraphs}
\end{figure}
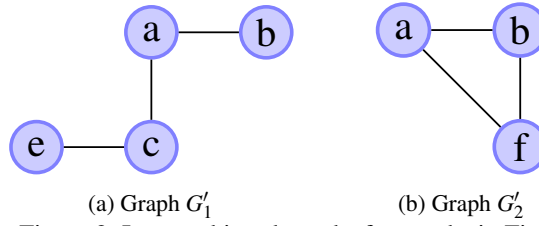

\begin{example}
    The graph $G'_1$ of Figure~\ref{fig:subgraphs} is an isomorphic subgraph of graphs $G_1$ and $G_2$ of Figure~\ref{fig:graphs}; it is not isomorphic to the graph $G_3$. The graph $G'_2$ is isomorphic to graphs $G_2$ and $G_3$, but it is not isomorphic to $G_1$.  \qed
\end{example}
\revision{In the general setting, the problem of deciding whether a subgraph isomorphism exists  (which is at the core of graph mining problems) is NP-complete \cite{DBLP:conf/stoc/Cook71}. However, several restricted settings have been identified, for which the problem can be solved in polynomial time, e.g., unique-labelled undirected graphs \cite{kimelfeld14complexity}.}


\subsection{Condensed Pattern Representations under Constraints}
In data mining, constraints are typically specified by the user to encode domain background knowledge. \citeN{DBLP:conf/cpaior/NegrevergneG15} distinguish four types of constraints: 1) constraints over the pattern (e.g., restriction on its 
size), 2) constraints over the cover set (e.g., minimal frequency), 3) constraints over the inclusion relation (e.g., maximal allowed gap in sequential patterns) and 4) constraints over the solution set (e.g., condensed representations).  

Orthogonally, constraints can be classified into \emph{local} and \emph{global} ones.  
A constraint is \emph{local} if deciding whether a given pattern satisfies it is possible without looking at other patterns. For example, minimal frequency or maximal pattern 
size are local constraints. On the contrary, deciding whether a pattern satisfies a \emph{global} constraint requires comparing it to other patterns. All constraints from the 4th group are global ones. 

As argued in Section~\ref{sec:intro}, the order in which constraints are applied influences the solution set \cite{DBLP:journals/kais/BonchiL06}. 
Following \citeN{DBLP:journals/kais/BonchiL06}, in this work we 
apply 
global constraints only after local ones.

We now present the notions required in our pattern mining framework. Here, the definitions are given for itemsets; for sequences and graphs they are identical up to substitution of $\subset$ with $\sqsubset$ (subsequence/subgraph relation). First, to rule out patterns that do not satisfy some of the local constraints, we introduce the notion of validity.

\begin{definition}[Valid pattern under constraints]\label{def:val}
    Let $C$ be a constraint function \revision{(local constraint)} from \patternspace to $\{ \top, \bot \}$ and let $p$ be a pattern in \patternspace. Then the pattern $p$ is called \emph{valid} iff $C(p) = \top$; otherwise it is referred \revision{to} as \emph{invalid}.
\end{definition}

\begin{example}\label{ex:valid}
    Let $C$ be a constraint function checking whether a given pattern is of size 
    at least 2. Then in Example~\ref{ex:it}, we have $C(\alpha_i)=\bot$, $i\in\{1,2,3\}$, and $C(\alpha_j)=\top$, $j\in\{4,5\}$. \qed
\end{example}

For detecting patterns that satisfy a given global constraint, the notion of \emph{dominance} is of crucial importance. Intuitively, a dominance relation reflects pairwise preference ($\subpattern$)
between patterns, and it is specific for each mining setting.  In this work we primarily focus on global constraints related to maximal, closed, free and skyline condensed representations, for which $\subpattern$ is defined as follows:

\begin{itemize}
    \item[(i)] \textbf{Maximal.} For itemsets $p$ and $q$, $p \subpattern q$ holds iff $p \subset q$
    \item[(ii)] \textbf{Closed.} For itemsets $p$ and $q$, $p \subpattern q$ holds iff $p \subset q$ and  $\mi{sup}(p) = \mi{sup}(q)$ 
    \item[(iii)] \textbf{Free.} For itemsets $p$ and $q$, $p \subpattern q$ holds iff $q \subset p$ and $\mi{sup}(p) = \mi{sup}(q)$ 
    \item[(iv)] \textbf{Skyline.} For itemsets $p$ and $q$, $p \subpattern q$ holds iff 
        \begin{itemize}
            \item[(a)] $\mi{sup}(p) \leq \mi{sup}(q)$ and $\size(p) < \size(q)$ or 
            \item[(b)] $\mi{sup}(p) < \mi{sup}(q) $ and $\size(p) \leq \size(q)$
        \end{itemize}
\end{itemize}

We are now ready to define dominated patterns under constraints.

\begin{definition}[Dominated pattern under constraints]\label{def:dom}
    Let $C$ be a constraint function, and let $p$ be a pattern, then $p$ is called \emph{dominated} iff there exists a pattern $p' \in \patternspace$ such that $p \subpattern p'$, and $p'$ is valid under $C$.
\end{definition}

\begin{example}
    In Example~\ref{ex:it} for the maximality constraint we have that $\alpha_1$ is dominated by $\alpha_4$, $\alpha_2$ by $\alpha_5$, while $\alpha_3$ both by $\alpha_4$ and $\alpha_5$. \qed
\end{example}

Exploiting the above definitions we obtain condensed patterns under constraints.

\begin{definition}[Condensed pattern under constraints]\label{def:con}
    Let $p$ be a pattern from \patternspace, and let $C$ be a constraint function, then a pattern $p$ is called \emph{condensed} under constraints iff it is valid and not dominated under $C$. 
\end{definition}
 \begin{example}
     For the constraint function selecting maximal itemsets of size at 
least $2$ and 
     \revision{support} at least $2$, $\alpha_4$ and $\alpha_5$ from Example~\ref{ex:it} are condensed patterns. \revision{The restriction on the pattern size rules out $\alpha_3$.} \qed
 \end{example}

\revision{Intuitively, a condensed representation is the smallest set of ``good'' patterns describing the data, i.e., a set that does not contain 
any redundant or invalid patterns. By redundant here, we mean patterns dominated 
by  others. 
More formally, }

\begin{definition}[Condensed representation]\label{def:condensed_representation}
    \revision{A \emph{condensed representation} is a set of all condensed patterns.}
\end{definition}
\revision{
Condensed representations 
allow one to get insight about the data without analyzing all frequent patterns. This is advantageous, since  typically there are orders of magnitude fewer condensed patterns than frequent ones. Moreover, many standard condensed representations (e.g., closed) allow a full reconstruction of the whole set of frequent patterns. 
In other words, from a substantially smaller set of condensed patterns, often the same knowledge can be extracted about the data, as from all frequent patterns.}
\revision{Note that in the literature condensed representations are also known under other names, e.g., 
\textit{dominating sets} \cite{dp2013}.}


\subsection{Answer Set Programming}

Answer Set Programming (ASP) \cite{GL1988} 
is a declarative problem solving paradigm oriented towards difficult search problems. ASP has its roots in Logic Programming and Nonmonotonic Reasoning.
An \emph{ASP program} $\Pi$ is a set of rules \revision{$r$} of the form
\begin{equation}
    \label{eq:asp:rule}
    \texttt{a\_0 :- b\_1, ..., b\_k, not b\_{k+1}, ..., not b\_m.}
\end{equation}
where $0 \leq k \leq m $,  \texttt{a\_0, b\_1, ..., b\_m} are classical literals, and \,{$\naf$}\,is 
default negation. The right-hand side of $r$ is its body, $\mi{body(r)}$, while the left-hand side is the head, $\mi{head(r)}$.  $\mi{body^+(r)}$ and $\mi{body^-(r)}$ \revision{stand for the positive and negative sets of atoms (respectively) that compose} $\mi{body(r)}$. A rule of the form \eqref{eq:asp:rule} is 
a \emph{fact} if $m=0$. We omit the symbol \texttt{:-} when referring to facts.  A rule without head literals is a \emph{constraint}. Moreover, a rule is \emph{positive} if $k=m$.

An ASP program $\Pi$ is \emph{ground} if it consists of only ground rules, i.e. rules without
variables. Ground instantiation $\mi{Gr}(\Pi)$ of a nonground program $\Pi$ is obtained by substituting variables with constants in all possible ways. The \emph{Herbrand universe}  $\mi{HU}(\Pi)$ (resp. \emph{Herbrand base} $\mi{HB}(\Pi)$) of $\Pi$, is the set of all constants occurring in $\Pi$, (resp. the set of all possible ground atoms that can be formed with predicates and constants appearing in $\Pi$). Any subset of $\mi{HB}(P)$ is a \emph{Herbrand interpretation}. ${\mi{MM}(\Pi)}$ denotes the subset-minimal Herbrand interpretation \revision{that is a \emph{model} of a ground positive program $\Pi$}.

The semantics of an ASP program is given in terms of its answer sets. An interpretation $A$ of $\Pi$ is an \emph{answer set} (or \emph{stable model}) of $\Pi$ iff $A = \mi{MM}(\Pi^A)$, where $\Pi^A$ is the \emph{Gelfond--Lifschitz (GL) reduct} \cite{GL1988} of $\Pi$, obtained from $Gr(\Pi)$ by removing (i) each rule $r$ such that $\mi{body}^-(r) \cap A\neq\emptyset$, and (ii) all the negative atoms from the remaining rules. The set of answer sets of a program $\Pi$ is denoted by $AS(\Pi)$.

\begin{example}
    Consider the program $\Pi$ given as follows:
    \smallskip

    \small{\leftline{
        $\renewcommand{\arraystretch}{1.1}
    \begin{array}{@{\,}l@{~~}l@{}}
        \texttt{\mbox{(1) }pattern(1); \mbox{(2) }pattern(2); \mbox{(3) }item(1,a);} \\ \texttt{\mbox{(4) }item(1,b); \mbox{(5) }item(2,a);} \\
        \texttt{\mbox{(6) }\revision{not\_superset(J,I)}:-pattern(I), item(I,V), I != J,} \\
        \phantom{\texttt{\mbox{(6) }not\_superset(J,I):-}}\texttt{pattern(J), not\ item(J,V).}\\
    \end{array}$}}
\smallskip

\normalsize{\revision{Apart from the facts (1)-(5) the program $\Pi$ contains a rule, which intuitively states that the pattern $\texttt{J}$ is not a superset of the pattern $\texttt{I}$ if $\texttt{I}$ has an item $\texttt{V}$ that $J$ does not have.
The grounding $\mi{Gr}(\Pi)$ of $\Pi$ is obtained from $\Pi$ by substituting \texttt{I,J,V} with the constants \texttt{1,2,a,b} in all possible ways. 
Consider the interpretation $A=\{\texttt{pattern(1),pattern(2),item(1,a),item(1,b),}$\\$\texttt{item(2,a),not\_superset(2,1)}\}$. The GL-reduct $\Pi^{A}(\Pi)$ for $A$ contains the facts (1)-() and the rule (6) with only positive atoms in its body, and with \texttt{I,J,V} substituted by \texttt{2,1,b} respectively.
$\mi{A}$ is the 
    minimal model of $\Pi^{A}(\Pi)$, and thus it is in $\mi{AS}(\Pi)$. \qed}}
\end{example}

\leanparagraph{\revision{Cardinality constraints}}\revision{ \emph{Cardinality constraints}  are extended literals \cite{DBLP:journals/ai/SimonsNS02}. They are of the form 
\begin{center}
\texttt{l$\{$b\_1, $\dotsc$, b\_m$\}$u}, 
\end{center}
for $m \geq 1$, where \texttt{l, u} are
lower and upper bounds on the cardinality of subsets of \texttt{$\{$b\_1, $\dotsc$, b\_m$\}$} satisfied in an encompassing answer set. They can appear in the head or in the body of a rule. A cardinality
constraint is satisfied in an answer set $A$, if the number of atoms from \texttt{b\_1, $\dotsc$, b\_m} belonging to $A$ is between \texttt{l} and \texttt{u}. 
}
\begin{example}
\revision{For instance, \texttt{1 \{a(X),b(X)\} 3} is satisfied in $A$, whenever between $1$ and $3$ instances of \texttt{a(X),b(X)} are true in $A$. \qed} 
\end{example}

\revision{Other relevant language constructs include \emph{conditional literals}. 
A \emph{conditional literal} is an expression of the form 
\begin{center}
$\{$\texttt{a:b\_1,...,b\_m}$\}$,
\end{center}
 where \texttt{a} and \texttt{b\_i} are possibly default negated literals. This expression denotes the set of atoms \texttt{a(X)} for which it holds that
\texttt{b\_1(X),...,b\_m(X)} are true. }
\begin{example}\revision{For example, the cardinality atom: \texttt{k \{in\_subset(X) : in\_set(X)\} k } expresses the condition that a subset has exactly \texttt{k} elements when the
predicate \texttt{in\_set(X)} defines the elements that belong to the set and the predicate
\texttt{in\_subset(X)} defines the subset. \qed}
\end{example}

\leanparagraph{\revision{Aggregate functions and aggregate atoms}} \revision{An \emph{aggregate function}  is of the form \texttt{f(S)}, where \texttt{S} is a set and \texttt{f} is a function name among \texttt{\#count, \#min, \#max, \#sum,} and \texttt{\#times}. 
An \emph{aggregate atom} is 
\begin{center}\texttt{Lg} $\prec_1$ \texttt{f(S)} $\prec_2$ \texttt{Rg},
\end{center} where \texttt{f(S)} is an \emph{aggregate function}, $\prec_1,\prec_2\in \{=,<,\leq,>,\geq\}$, and \texttt{Lg} and \texttt{Rg} (called left guard and right guard, respectively) are terms. One of \texttt{Lg}$\prec_1$ and $\prec_2$\texttt{Rg}  can be omitted, in which case, ``$0$'' and ``$+\infty$'' are assumed, respectively. If both $\prec_1$ and $\prec_2$ are present, we assume for simplicity that $\prec_1 \in \{<,\leq \}$ if and only if $\prec_2\in\{<,\leq\}$, and that both $\prec_1$ and $\prec_2$ are different from $=$ \cite{dlv_aggregates}.}
\revision{We consider the standard semantics of aggregates:
\begin{itemize}
    \item \texttt{\#count($X$)}, defined over a multiset of atoms $X$, is the number of atoms $X$ that hold in an answer set (zero for the empty set)
    \item \texttt{\#min($X$)},   defined over a multiset of atoms $X$, is a minimal atom in $X$ that holds in an answer set
    \item \texttt{\#max($X$)},   defined over a multiset of atoms $X$, is a maximal atom in $X$ that holds in an answer set
    \item \texttt{\#sum($X$)},   defined over a multiset of atoms $X$, is the sum of atoms $X$ (which are typically numbers) that hold in an answer set
    \item \texttt{\#time($X$)},  defined over a multiset of atoms $X$, is the product of atoms $X$ (which are typically numbers) that hold in an answer set (one for the empty set).
\end{itemize}}

\leanparagraph{\revision{Encoding methodology and ASP solvers}}\revision{In this paper we make use of two existing ASP systems Clasp \cite{clasp} and WASP \cite{wasp}. We mostly focus on the former and use the 
latter to provide an experimental comparison on how two different systems  
perform on various pattern mining tasks with respect to runtime and memory usage.}

\revision{Clasp extends the standard ASP theory described here with a number of features such as incremental grounding, preferences, Satisfiability Modulo Theories, various solving parameters such as brave and cautious reasoning, parallel executions, heuristics and meta-programming \cite{ASPbook}.}

\revision{A typical modelling approach in ASP 
  follows the guess and check paradigm, where at first we define potential stable model candidates 
(typically through non-deterministic constructs) and then eliminate invalid candidates (typically through integrity constraints) \cite{clasp,ASPbook,whatisasp}, which in a nutshell allows us to write the following formula for ASP modelling:}

\begin{center}
    {\bfseries ASP program = Data + Generator + Tester ( + Optimizer)} \footnote{\url{https://www.cs.uni-potsdam.de/\textasciitilde torsten/Potassco/Slides/asp.pdf}}
\end{center}

\revision{We follow this modelling paradigm throughout the paper. To demonstrate it, consider as an example, $n$-queen modelling problem, where one needs to put a number of queens on the board such that no queen attacks another. This can be modelled as in Listing \ref{lst:n_queens_example_modelling}. First we specify data as facts, in Lines \ref{background:line:data1} and \ref{background:line:data2}, then we define the set of possible answer sets using a choice rule in Line \ref{background:line:choice} and finally we validate them in Lines \ref{background:line:int1},\ref{background:line:int2} and \ref{background:line:int3}.}

\begin{lstlisting}[language=ASPlang,escapeinside={@}{@},label=lst:n_queens_example_modelling,caption=An ASP guess-and-check paradigm modelling example on the $n$-queen problem]
@\commenttextasp{\% Data specifying the board}@
col(1..k).@\label{background:line:data1}@
row(1..k).@\label{background:line:data2}@
@\commenttextasp{\% Choice rule to specify possible stable models}@
k { queen(Row,Col) : col(Col), row(Row) } k. @\label{background:line:choice}@
@\commenttextasp{\% Integrity constraint to validate the candidate models}@
:- queen(Rw,Cw), queen(Rb,Cb), Rw  = Rb, Cw != Cb. @\label{background:line:int1}@
:- queen(Rw,Cw), queen(Rb,Cb), Rw != Rb, Cw  = Cb. @\label{background:line:int2}@
:- queen(Rw,Cw), queen(Rb,Cb), Rw != Rb, | Rw - Rb | = | Cw - Cb |.@\label{background:line:int3}@
\end{lstlisting}



\section{Hybrid ASP-based Mining Approach}\label{sec:problem}

In this section we present our hybrid method for frequent pattern mining. Unlike previous ASP-based mining methods, our approach 
consists of \revision{two steps, where in the first step, we apply highly optimized algorithms for frequent pattern discovery and, in the second step, we use a declarative ASP solver for their convenient post-processing. }
Here, we mainly focus on itemsets, sequence and graph mining. 

Given a frequency threshold $\sigma$, a dataset $D$ and a set of constraints $\cC=\cC_l\cup\cC_g$, where $\cC_l$ and $\cC_g$ are respectively local and global constraints, we proceed in two steps as follows. 
\bigskip

\leanparagraph{Step 1} First, we launch a dedicated optimized algorithm to extract all frequent patterns from a given dataset, satisfying the minimal frequency threshold $\sigma$. Here, any frequent pattern mining algorithm can be invoked. 
We use Eclat \cite{eclat} for itemsets, PPIC \cite{PPIC} for sequences and gSpan \cite{gspan} for graphs.
\bigskip

\leanparagraph{Step 2} Second, the computed patterns are post-processed using the declarative means to select a set of \emph{valid} patterns (i.e., those satisfying constraints in $\cC_l$). For that the frequent patterns obtained in Step 1 are encoded as facts \texttt{item(i,j)} for itemsets and \texttt{seq(i,j,p)} for sequences, where \texttt{i} is the pattern's ID, \texttt{j} is an item contained in it and \texttt{p} is its position. \revision{Analogously, we encode graphs with unique labels using facts \texttt{graph\_u(i,j,k)}, where \texttt{i} is the pattern's ID, \texttt{j} and \texttt{k} are node labels, and the respective fact means that in the graph \texttt{i}, the node \texttt{j} is connected to \texttt{k}. In the general setting, we have \texttt{graph(i,j,k,l)}, where \texttt{i} is again the pattern's ID, \texttt{j} and \texttt{k} are node labels, and \texttt{l} is the label of an edge connecting \texttt{j} and \texttt{k} in \texttt{i}.} 
The local constraints in $\cC_l$ are represented as ASP rules, which collect IDs of patterns satisfying constraints from $\cC_l$ into the dedicated predicate \texttt{valid}, while the rest of the IDs are put into the \texttt{not\_valid} predicate. 

Finally, from all valid patterns a desired condensed representation is constructed by storing patterns \texttt{i} in the \texttt{selected} predicate if they are not \texttt{dominated} by other valid patterns based on constraints from $\cC_g$. Following the principle of \cite{DBLP:conf/lpnmr/Jarvisalo11}, for itemsets and sequences in our work every answer set represents a single desired pattern, which satisfies both local and global constraints (for graphs slight variations apply, as we discuss later in this section). The set of all such patterns forms a condensed representation. In what follows we present our encodings of local and global constraints in details. 

\subsection{Encoding Local Constraints} 

In our declarative program we specify local constraints by the predicate \texttt{valid}, which reflects the conditions given in Definition~\ref{def:val}. For every constraint in $\cC_l$ we have a set of dedicated rules, stating when a pattern is not valid. 
For instance, a constraint checking whether the cost of items in a pattern exceeds a given threshold $N$ is encoded as 

\small{\begin{center}
\texttt{not\_valid(I) :- \#sum\{C,\revision{J}:item(I,J),cost(J,C)\} > N, pattern(I).}
\end{center}}

\normalsize{A similar rule for sequences can be defined as follows: }

\small{\begin{center}
\texttt{not\_valid(I) :- \#sum\{C,\revision{J,P}:seq(I,J,P),cost(J,C)\} > N, pattern(I).}
\end{center}}

\normalsize{Analogously, one can specify arbitrary domain constraints on patterns. }

\begin{example}
Consider a dataset storing moving habits of young people during their studies. Let the dedicated frequent sequence mining algorithm return the following patterns: $S_1=\mi{\tuple{bG\;\revision{mA}\;ba\;mG\;ma}}$; $S_2=\mi{\tuple{\revision{bA}\;mG\;ba\;\revision{mA}\;ma}}$; $S_3=\mi{\tuple{\revision{bUS\;mA\;ba\;mUS\;ma}}}$, where $\mi{bG,\revision{bA,bUS}}$ stand for born in Germany, \revision{Austria} and \revision{US}, $\mi{ba,ma}$ stand for bachelors and masters and the predicates $\mi{\revision{mG,mA, mUS}}$ reflect that a person moved to Germany, \revision{Austria and US}, respectively. Suppose, we are only interested in moving habits of \revision{German native speakers}, who got their masters degree from a German university. The local domain constraint expressing this would state that (1) \revision{$\mi{bUS}$} should not be in the pattern, while (2) either both $\mi{bG}$ and $\mi{ma}$ should be in it without any \revision{$\mi{mA}$ or $\mi{mUS}$} in between or $\mi{mG}$ should \revision{directly} precede $\mi{ma}$. These constraints are encoded in the program in Listing~\ref{ex:move}. From the answer set of this program we get that both  $S_2$ and $S_3$ are not valid, while $S_1$ is. \qed

 \begin{figure}[t]
 \small{
 \begin{lstlisting}[language=ASPlang,label=ex:move,caption=Moving habits of people during studies, escapeinside={@}{@}]
time(1..5).
@\commenttextasp{\% people born in Germany or \revision{Austria} are \revision{German-speaking}}@
@\revision{gs(I) :- seq(I,bG,P).}@
@\revision{gs(I) :- seq(I,bA,P).}@
@\commenttextasp{\% collect those who moved to \revision{Austria or US} before P}@
moved_before(X,P) :- seq(X,@\revision{mA,}@P1), P > P1, time(P), time(P1).
moved_before(X,P) :- seq(X,@\revision{mUS,}@P1), P > P1, time(P), time(P1)
@\commenttextasp{\% collect those who moved to \revision{Austria or US} after P and before masters}@
moved_after(X,P) :- seq(X,@\revision{mA,}@P1), seq(X,ma,P2), P < P1, 
                    P1 < P2, time(P), time(P1), time(P2).
moved_after(X,P) :- seq(X,@\revision{mUS,}@P1), seq(X,ma,P2), P < P1, 
                    P1 < P2, time(P), time(P1), time(P2).
@\commenttextasp{\% keep \revision{German speakers} who moved to Germany straight before masters}@
keep(X) :- seq(X,ma,P+1), seq(X,mG,P), @\revision{gs(X).}@
@\commenttextasp{\% keep Germans who did not move before masters}@
keep(X) :- seq(X,bG,P1), seq(X,ma,P), not moved_before(X,P).
@\commenttextasp{\% keep German \revision{speakers} whose last move before masters was to Germany}@
keep(X) :- seq(X,mG,P1), seq(X,ma,P2), P1 < P2,  
           @\revision{gs(X),}@not moved_after(X,P1).
@\commenttextasp{\% a pattern is not valid, if it should not be kept}@
not_valid(X) :- pattern(X), not keep(X).
 \end{lstlisting}
}
 \end{figure}
\end{example}

\normalsize{
To combine all local constraints from $\cC_l$ we add to a program a generic rule specifying that a pattern \texttt{I} is valid whenever \texttt{not\_valid(I)} cannot be inferred. }

\small{
\begin{center}
\texttt{valid(I) :- pattern(I), not not\_valid(I)}
\end{center}}

\normalsize{Patterns \texttt{i}, for which \texttt{valid(i)} is deduced are then further analyzed to construct a condensed representation based on global constraints from $\cC_g$.}

\subsection{Encoding Global Constraints}

The key for encoding global constraints is the declarative formalization of the dominance relation (Defintion~\ref{def:dom}). For example, for itemsets the maximality constraint boils down to pairwise checking of subset inclusion between patterns. For sequences this requires a check of embedding existence between sequences.

 Regardless of a pattern type from $\cL$ 
and a constraint from $\cC_g$ every encoding presented 
in this section is supplied with a rule, which guesses (\texttt{selected/1} predicate) a single \texttt{valid} pattern to be a candidate for inclusion in the condensed representation, and a constraint that rules out \texttt{dominated} patterns thus enforcing a different guess. 

\small{\begin{center}
\texttt{1 \{selected(I) : valid(I)\} 1.}
\end{center}

\begin{center}
\texttt{ :- dominated.}
\end{center}}


 \begin{figure}[t]
\small{
 \begin{lstlisting}[language=ASPlang,label=lst:max,caption=Maximal itemsets encoding, escapeinside={@}{@}]
@\commenttextasp{\% J is not a superset of I if I has items that are not in J}@
not_superset(J) :- selected(I), item(I,V), not item(J,V), 
                 valid(J), I != J.
@\commenttextasp{\% derive dominated whenever I is a subset of J}@
dominated :- selected(I), valid(J), 
             I != J, not not_superset(J).
\end{lstlisting}}
 \end{figure}
 \normalsize{In what follows, we discuss concrete realizations of the dominance relation both for itemsets and sequences for various global constraints, i.e., we present specific rules related to the derivation of the \texttt{dominated/0} predicate.
}

\leanparagraph{Itemset Mining} We first provide an encoding for maximal itemset mining in Listing~\ref{lst:max}. To recall, a pattern is \emph{maximal} if none of its supersets is frequent. An itemset $I$ is included in $J$ iff for every item $i\in I$ we have  $i\in J$. We encode the violation of this condition in lines (1)--(3). The second rule presents the dominance criteria.
\revision{We provide a correctness proof of this encoding in Appendix. The correctness of the rest of the encodings can be analagously shown.}

For closed itemset mining a simple modification of Listing~\ref{lst:max} is required. An itemset is \emph{closed} if none of its supersets has the same support. Thus to both of the rules from Listing~\ref{lst:max} we need to add atoms \texttt{support(I,X), support(J,X)}, which store the support of \texttt{I} and \texttt{J} respectively (extracted from the output of Step 1).  


For free itemset mining the rules of the maximal encoding are changed as follows:
\medskip

\small{\begin{lstlisting}[language=ASPlang,firstnumber=4,label=lst:free,escapeinside={@}{@}]
not_superset(J) :- selected(I), item(J,V), not item(I,V), 
                   valid(J), I != J.
dominated :- selected(I), valid(J), support(I,X),
             I != J, not not_superset(J), support(J,X).
\end{lstlisting}}

\begin{figure}[t]
\small{\begin{lstlisting}[language=ASPlang,label=lst:sky,caption=Skyline itemsets encoding, escapeinside={@}{@}]
@\commenttextasp{\% support and size comparison among patterns}@
g_size_geq_fr(J) :- selected(I), valid(J), support(I,X), 
                    support(J,Y), size(I,Si), size(J,Sj), 
                    Si <  Sj, X <= Y. 
geq_size_g_fr(J) :- selected(I), valid(J), support(I,X), 
                    support(J,Y), size(I,Si), size(J,Sj), 
                    Si <= Sj, X < Y. 
@\commenttextasp{\% derivation of the domination condition}@
dominated :- valid(J), g_size_geq_fr(J). 
dominated :- valid(J), geq_size_g_fr(J). 
\end{lstlisting}}
\end{figure}
\normalsize{Finally, the skyline itemset 
encoding is given in Listing~\ref{lst:sky}, where the first two rules specify the conditions (a) and (b) for skyline itemsets as specified in Section~\ref{sec:prelim}. }

\begin{figure}[t]
\small{ \begin{lstlisting}[language=ASPlang,label=lst:maxs,caption=Maximal sequence encoding, escapeinside={@}{@}]
@\commenttextasp{\% if V appears in a valid pattern I, derive in(V,I)}@
in(V,I) :- seq(I,V,P), valid(I).
@\commenttextasp{\% J is not a superset of I if I has V that J does not have}@
not_superset(J) :- selected(I), valid(J), I != J, 
                   seq(I,V,P), not in(V,J).
@\commenttextasp{\% if for a subseq <V,W> in I there is V followed}@
@\commenttextasp{\% by W in J then deduce domcand(V,J)}@
domcand(V,J,P) :- selected(I), seq(I,V,P), seq(I,W,P+1), I != J
                  valid(J), seq(J,V,Q), seq(J,W,Q'), Q'>Q.
@\commenttextasp{\% if domcand(V,J) does not hold for some V in I}@ 
@\commenttextasp{\% and a pattern J then derive not\_dominated\_by(J)}@
not_dominated_by(J) :- selected(I), seq(I,V,P), seq(I,W,P+1), 
                       I != J, valid(J), not @\revision{domcand(V,J,P)}.@
@\commenttextasp{\% if neither not\_dominated\_by(J) nor not\_superset(J)}@
@\commenttextasp{\% are derived for some J, then I is dominated}@
dominated :- selected(I), valid(J), I != J, 
             not not_superset(J), not not_dominated_by(J). \end{lstlisting}}
 \end{figure}

\normalsize{\leanparagraph{Sequence Mining} 
The subpattern relation for sequences is slightly more involved, than for itemsets, as it preserves the order of elements in a pattern. To recall, a sequence $S$ is included in $S'$ iff an embedding $e$ exists, such that $S \sqsubseteq_e S'$.}

In Listing~\ref{lst:maxs} we present the encoding for maximal sequence mining. A selected pattern is not maximal if it has at least one valid superpattern. We rule out patterns that are for sure not superpatterns of a selected sequence. First, \texttt{J} is not a superpattern of \texttt{I} if \revision{it is not a superset of \texttt{I}} (lines (4)--(5)), i.e., if \revision{\texttt{not\_superset(J)}} is derived, then \texttt{J} does not dominate \texttt{I}. If \texttt{J} is a superset of \texttt{I} then to ensure that \texttt{I} is not dominated by \texttt{J}, the embedding existence has to be checked (lines (6)--(9)). \texttt{I} is not dominated by \texttt{J} if an item exists in \texttt{I}, which together with its sequential neighbor cannot be embedded in \texttt{J}.  This condition is checked in lines (10)--(13), where \texttt{domcand(V,J,P)} is derived if for an item \texttt{V} at position \texttt{P} and its follower, embedding in \texttt{J} can be found. 

The encoding for closed sequence mining is obtained from the maximal sequence encoding analogously as it is done for itemsets. The rules for free sequence mining are constructed by substituting lines (4)--(13) of Listing~\ref{lst:maxs} with the following ones:

\small{\begin{lstlisting}[language=ASPlang,firstnumber=4,label=lst:free,escapeinside={@}{@}]
not_superset(J) :- selected(I), in(V,J), 
                   not in(V,I), I != J.
@\revision{domcand(V,J,P)}@:- selected(I), seq(J,V,P), @\revision{seq(J,W,P+1),}@ 
                @\revision{seq(I,V,Q),}@seq(I,W,Q'), Q'>Q, I != J.
not_dominated_by(J) :- selected(I), valid(J), I != J, 
                       seq(J,V,P), seq(J,W,P+1), 
                       not @\revision{domcand(V,J,P).}@ \end{lstlisting}}

\normalsize{Finally, 
the encoding for mining skyline sequences coincides with the skyline itemsets encoding, which is provided in Listing~\ref{lst:sky}.

\normalsize{\leanparagraph{Graph Mining}} Graphs represent the most complex pattern type. The subpattern relation between graphs amounts to testing subgraph isomorphism between them. In general, this problem is \NP-complete; however, for some restricted graph types, it is solvable in polynomial time.

\paragraph{\revision{Uniquely Labelled Graphs.}} One of such graph types are undirected graphs, where every node has a unique label (and consequently, every edge $(u, v)$ is labelled uniquely as $(l(u),l(v))$). For instance, the first two graphs in Figure~\ref{fig:graphs} fall into this category, while the third one does not. For this restricted graph type, we have that $G$ is subgraph isomorphic to $G'$ iff every edge in $G$ is also present in $G'$. Therefore, the subgraph isomorphism test for these restricted graphs essentially boils down to subpattern test for itemsets \cite{neumann17reductions}.

We treat every edge in a graph as an item. Since the graph is allowed to have only unique labels, there cannot be any repetitions of the edge labels. The encoding of maximal frequent graph patterns is presented in Listing~\ref{lst:maxgr}. A selected graph pattern is not maximal if at least one of its frequent supergraphs is valid. Similar to the case of itemsets and sequences we rule out the patterns that are guaranteed to be not maximal. A pattern \texttt{J} is not a superpattern of \texttt{I} if \texttt{I} contains an edge that \texttt{J} does not have.  

\begin{figure}[t]
    \small{\begin{lstlisting}[language=ASPlang,firstnumber=4,label=lst:maxgr,caption=Maximal graph encoding (unique labeled; \revision{IDs represent labels as well; \doublerev{graph\_u(I,X,Y)} represents here a graph I with an edge between X and Y, whose values are their labels respectively)}, escapeinside={@}{@}]
@\commenttextasp{\% I is not a subgraph of J if I has an edge (X,Y) that J does not have}@
not_supergraph(J) :- selected(I), @\revision{graph\_u(I,X,Y),}@ valid(J),
                   not @\revision{graph\_u(J,X,Y),}@ not @\revision{graph\_u(J,Y,X),}@ I != J.
@\commenttextasp{\% derive dominated, whenever I is a subgraph of J}@
dominated :- selected(I), valid(J), not not_supergraph(J), I != J.
\end{lstlisting}}
\end{figure}

The encoding for closed frequent graph patterns differs from the one for maximal graph patterns only in that in the second rule in Listing~\ref{lst:maxgr} the atoms \texttt{support(I,X), support(J,X)} are added. Encodings for other condensed representations are analogous.

\begin{example}
The edges of the graph $G_1$ in Figure~\ref{fig:graphs} are represented with the following facts \texttt{graph\_u(g1,e,c), graph\_u(g1,c,a)}, etc. Given the unique-labeled graphs $G_1$ and $G_2$ and the frequency threshold $\sigma=2$, we have that both \texttt{\{(a,b)\}} and \texttt{\{(e,c),(c,a),(a,b)\}} are frequent subgraphs. However, only the latter graph pattern is maximal. \qed
\end{example}

\paragraph{\revision{General Case of Graph Mining.}} The encoding for maximal (closed, etc) graph mining problem in the general case is slightly more complicated. For the maximal constraint we depict the encoding in Listing~\ref{lst:graph_mining_maximal_general} (the rest of the constraints are treated analogously). Since the subgraph isomorphism check is an NP-complete problem, after obtaining a set $F$  of frequent candidate graph patterns using a dedicated algorithm, we perform a dominance check using the ASP program in Listing~\ref{lst:graph_mining_maximal_general} for each pattern separately (unlike in earlier presented encodings, where such a check was done for all frequent patterns jointly within a single ASP program). 

More specifically, the solver receives as input a selected graph candidate $G$ and the rest of the frequent patterns in $F$ excluding $G$. 
The graph $G$ is represented using two predicates: \texttt{selected\_node(v,lv)} reflecting labelled vertices of $G$ and \texttt{selected\_edge(v,w,le)}, where \texttt{v,w} are nodes and the edge \texttt{(v,w)} is labeled with \texttt{le} in $G$. First, in (2) of Listing~\ref{lst:graph_mining_maximal_general} the guess is performed on a graph in the set $F$ of frequent patterns, to which $G$ can be mapped, i.e., 
a dominating candidate pattern is guessed. Second, in (4) 
the guess on a mapping from 
$G$ to the dominating candidate graph 
is done. Finally, in (6)-(9) the mapping is validated by 
two  integrity constraints. \revision{Finally, the constraint in (10)-(11) ensures the injectivity of the constructed mapping.} 
If 
the solver returns an answer set, then the selected graph $G$ is removed from the dataset as a dominated pattern, i.e., it is guaranteed to be not maximal.

\begin{figure}[t]
    \small{\begin{lstlisting}[language=ASPlang,label=lst:graph_mining_maximal_general,caption=Maximal graph encoding (core); general case \revision{with non-unique labels; the method picks a graph from the set of patterns and a) removes it from the graph dataset (we know the picked graph would always match itself) b) reify it as \texttt{selected\_node(X,L)} and \texttt{selected\_edge(X,Y,L)}, then the encoding is trying to construct an injective mapping \texttt{map(X,N)}, which maps nodes in the originally selected graph to the one picked in the dataset, i.e., \texttt{mapped\_to(I)} to one of the graphs in the dataset, i.e., the graph predicate (since it constructs a mapping names or ordering of nodes does not play a role here) such that it preserves edges and labels. If it fails to construct such a mapping, then the original selected graph is maximal}, escapeinside={@}{@}]
    @\commenttextasp{\% pick one graph to map to \revision{the originally selected}}@
1 { mapped_to(X) : graph(X,_,_,_)} 1.
@\commenttextasp{\% guess the mapping to the selected graph}@
1 { map(X,N) : node(I,N,_) } 1 :- selected_node(X,_), @\revision{ mapped\_to(I).}@
@\commenttextasp{\% check mapping validity on the edges with labels}@
:- mapped_to(I), map(X1,V1), map(X2,V2), 
   selected_edge(X1,X2,L),  not graph(I,V1,V2,L).
@\commenttextasp{\% check validity on the nodes}@
:- mapped_to(I), map(X,V), selected_node(X,L), not node(I,V,L).
@\commenttextasp{\% enforces injectivity of the mapping}@
:- map(X,V1), map(X,V2), V1 != V2.
\end{lstlisting}}
\end{figure}




\section{Covering with Approximate Patterns}\label{sec:tiling}

So far, we have concentrated on exact patterns, that is, patterns that are present exactly in the data; for example, an itemset is present only in transactions that contain every item in the itemset. This is a standard assumption in pattern mining, and it is heavily utilized by the pattern mining algorithms. But it also means that the patterns are not robust against noise: a single item missing from a single transaction can turn a frequent itemset into an infrequent one. To make the patterns more robust to noise, we can 
study approximate patterns, that is, patterns that are \revision{considered as} contained in transactions even if the transaction does not contain all of the items in the pattern. In this section, we demonstrate how our hybrid approach can be used also with approximate patterns.

The approximate patterns we are interested in are based on the concept of \emph{tiles} \cite{tiling}.
Let $\alpha$ be an itemset and $D$ be a database. The \emph{tile} corresponding to
$\alpha$ is defined as $\tau(\alpha,D)=\{(t_{id},i)\mid t_{id} \in D_{\alpha}, i\in \revision{\alpha}\}$. When $D$ is clear from the context, we write $\tau(\alpha)$.  The area of $\tau(\alpha)$ is equal to its cardinality.

\begin{example}
  \label{ex:tile}
The tile corresponding to the itemset $\{\mi{b,e}\}$ from Table~\ref{tab:it} is $\{\mi{(1,b),(2,b), (1,e), (2,e)}\}$.
\end{example}

A \emph{tiling} $T = \{\tau(\alpha_1),\dotsc, \tau(\alpha_k)\}$ consists of a finite number of tiles. Its area is $\mi{area}(T,D)=\abs{\tau(\alpha_1) \cup,\dotsc,\cup \tau(\alpha_k)}$. In \emph{Maximum $k$-Tiling} the goal is to find a tiling of $k$ tiles with the maximum area, whereas in the \emph{Minimum Tiling} the goal is to find the least number of tiles such that \revision{for all $t_{id}\in D$ and for all $i\in t_{id}$, there exists at least one tile $\tau(\alpha)$ such that $(t_{id}, i)\in \tau(\alpha)$} \cite{tiling}.

Another way to define tiles is via binary matrices. A transaction database $D$ can be regarded as a binary matrix $\mD$ of size
$m \times n$, where $m$ is the number of transactions in
$D$ and $n$ is the number of items in \revision{$\cI$}. The $(i, j)$th element in the binary matrix corresponding to
$D$ is equal to $1$ if the $i$th transaction contains item
$j$, and it is equal to $0$ otherwise. In this framework, a tile is a \emph{rank-1} matrix $\mT$ that is \emph{dominated} by the data matrix $\mD$. Matrix $\mT$ is rank-1 if it is an outer product of two binary vectors, that is, $\mT = \vu\vv^T$ where $\vu\in\B^{m}$ and $\vv\in\B^n$. Matrix $\mT=(t_{ij})$ is dominated by matrix $\mD=(d_{ij})$ if $t_{ij} \leq d_{ij}$ for all $i$ and $j$. 

\begin{example} The transaction database in Table~\ref{tab:it} corresponds to the following binary matrix:
\[
\begin{pmatrix}
    1  & 1 & 0 & 1 & 1 \\
    0  & 1 & 1 & 0 & 1 \\
    1  & 0 & 0 & 0 & 1  
\end{pmatrix}\; ,
\]
and the tile from Example~\ref{ex:tile} corresponds to the matrix
\[
  \begin{pmatrix}
    0  & 1 & 0 & 0 & 1 \\
    0  & 1 & 0 & 0 & 1 \\
    0  & 0 & 0 & 0 & 0  
  \end{pmatrix}
  =
  \begin{pmatrix}
    1 \\
    1 \\
    0
  \end{pmatrix}
  \begin{pmatrix}
    0 & 1 & 0 & 0 & 1
  \end{pmatrix}\; .
\]
\qed
\end{example}

A tiling is an element-wise logical disjunction of the matrices corresponding to the tiles,
\[
  \mS = \mT_1 \lor \mT_2 \lor \cdots \lor \mT_k\; .
\]
Assuming that \revision{$\mT_{\ell} = \vu_{\ell}\vv_{\ell}^T$,} we can write each element of $\mS=(s_{ij})$ as
\[
  s_{ij} = \bigvee_{\ell=1}^k u_{i\ell}v_{j\ell}\; ,
\]
that is, $\mS$ is the \emph{Boolean matrix product} of matrices $\mU = [\vu_1\; \vu_2\; \cdots \; \vu_k]$ and $\mV = [\vv_1\; \vv_2\; \cdots \; \vv_k]$ (see \citeN{miettinen09matrix} for more discussion on the connections between pattern mining and Boolean matrix factorization).

The formulation of tiles as rank-1 binary matrices immediately suggests the concept of \emph{approximate tiles} as \revision{possibly} non-dominated rank-1 matrices and \emph{approximate tiling} as approximate Boolean matrix factorization. We will call both exact and approximate tiles simply as tiles from now on.

\begin{example}\label{ex:tiling}
Consider the following Boolean matrix: 
\begin{center}
\begin{tikzpicture}
        \matrix [matrix of math nodes,left delimiter=(,right delimiter=)] (m)
        {
            1 &1 &0  \\               
            1 &0 &1  \\               
            0 &1 &1  \\           
        };  
        \draw[line width=0.45mm, color=red] (m-1-1.north west) -- (m-1-3.north east) -- (m-2-3.south east) -- (m-2-1.south west) -- (m-1-1.north west);
        \draw[line width=0.45mm, color=blue] (m-2-1.north west) -- (m-2-2.north east) -- (m-3-2.south east) -- (m-3-1.south west) -- (m-2-1.north west);
        \draw[line width=0.45mm, color=black] (m-2-2.north west) -- (m-2-3.north east) -- (m-3-3.south east) -- (m-3-2.south west) -- (m-2-2.north west);
    \end{tikzpicture}
\end{center}
The areas with the red, blue and black border highlight the respective three approximate tiles of the matrix. \qed
\end{example}

A common approach to calculate an approximate Boolean matrix factorization (or tiling) is to first calculate a set of \emph{candidate tiles} (or rank-1 matrices) and then select the final set from these \cite{miettinen08discrete,tyukin14bmad}. In what follows, we concentrate on a version of this problem where the user gives the target quality, and we try to find a set of tiles that obtains  this quality.

\begin{definition}[Approximate Tile Selection Problem]
  \label{def:probtile}
Given a binary matrix $\mD\in\B^{m\times n}$, a set $T$ of tiles (rank-1 matrices) and an integer threshold $\sigma$, find a subset $T' = \{\mT^{(1)}, \mT^{(2)}, \ldots, \mT^{(k)}\}\subseteq T$, such that when we write $\mS = (s_{ij})$ with $s_{ij} = t^{(1)}_{ij} \lor t^{(2)}_{ij} \lor \cdots \lor t^{(k)}_{ij}$,
the error
\begin{equation}\label{eq:probtile_err}
  \abs{\{(i, j)\mid d_{ij} = 1 \land s_{ij} = 0\}} + \abs{\{(i, j) \mid d_{ij} = 0 \land s_{ij} = 1\}}
\end{equation}
(i.e., the number of ones outside of the tiling plus the number of zeros inside the tiling) is smaller then the threshold $\sigma$.
\end{definition}

It is worth noticing that \eqref{eq:probtile_err} corresponds to the Hamming distance between $\mD$ and $\mS$. The decision version of the Approximate Tile Selection problem is \NP-hard \cite{miettinen15generalized}, and the optimization version (finding the smallest possible error) is \NP-hard to approximate to within $\Omega(2^{\log^{1-\epsilon}\abs{\mD}})$ for any $\epsilon > 0$, where $\abs{\mD}$ is the number of 1s in $\mD$ \cite{miettinen15generalized}. The problem has two alternative characterizations, either as a variant of the famous Set Cover problem called \emph{Positive-Negative Partial Set Cover} \cite{miettinen08positive-negative} or as a form of \emph{Boolean linear programming}: given a binary design matrix $\mA$, and a binary target vector $\vb$, find a binary vector $\vx$ that minimizes the Hamming distance between $\vb$ and $\mA\vx$, where the matrix-vector product is over Boolean algebra.

\begin{figure}[t]
\small{ \begin{lstlisting}[language=ASPlang,label=lst:tiles,caption=Encoding for the approximate tiling problem from Def.~\ref{def:probtile}, escapeinside={@}{@}]
@\commenttextasp{\% for every tile store its ID and coordinates of 1s that it encodes}@
tileid(X) :- tile(X,Y,Z).
one(Y,Z)  :- tile(X,Y,Z).
@\commenttextasp{\% \revision{among tiles guess at least one to be included in the tiling}}@
1 { @\revision{intiling(X)}@ : tileid(X)}.
@\commenttextasp{\% collect and count ones that are outside the constructed tiling}@
incell(Y,Z) :- intiling(X), tile(X,Y,Z).
outsideone(Y,Z) :- tile(X,Y,Z), not incell(Y,Z).
numberofoutsideones(N) :- N = #count{Y, Z : outsideone(Y,Z)}.
@\commenttextasp{\% collect and count zeros that are inside the constructed tiling}@
insidezero(Y,Z) :- intiling(X), relitem(X,Y), 
                   reltrans(X,Z), @\revision{not one(Y,Z).}@ 
numberofinsidezeros(N) :- N = #count{Y, Z : insidezero(Y,Z)}.
@\commenttextasp{\% compute the total error}@
er(0,K) :- @\revision{numberofinsidezeros(K).}@
er(1,K) :- @\revision{numberofoutsideones(K).}@
totalerror(N) :- N = #sum{X, Y : er(Y,X)}.
@\commenttextasp{\% ensure that the total error does not exceed the threshold}@
:- totalerror(N), threshold(K), N > K.
\end{lstlisting}}
\end{figure}

For computing a set of candidate tiles $T$ to choose from, effective implementations exist
(see, e.g., \citeN{tyukin14bmad}). From these candidates a subset needs to be selected, which will be a solution to the problem from Def.~\ref{def:probtile}. In Figure~\ref{lst:tiles} we provide an ASP program, whose answer sets exactly correspond to solutions to the above problem.

\revision{In the input the ASP program gets a set of tile candidates to choose from. 
A natural encoding of tiles is via the positions of 1s that they contain, i.e., for a tile with ID \texttt{id} that has 1 in the intersection of a column \texttt{i} and a row \texttt{t}, we could store a fact \texttt{tile(id,i,t)}. \doublerev{For instance, provided that the ID of the red tile from Example~\ref{ex:tiling} is \texttt{1}, it can be represented as a set of facts \texttt{tile(1,1,1),tile(1,2,1),tile(1,1,2),tile(1,3,2)}.} However, 
storing only information about 1s in a tile is insufficient for its encoding, \doublerev{since multiple tiles can have 1s in the same positions, 
as illustrated next.}}
\begin{example}\label{ex:tileenc}
\revision{\doublerev{Consider the matrix in Figure~\ref{fig:tileenc} and two of its tiles. If we only encode tiles using the positions of 1s in them, then both of the tiles in Figure~\ref{fig:tileenc} will have the exact same representation given by the following facts \texttt{tile(1,1,1), tile(1,4,1), tile(1,1,3), tile(1,4,3).}\qed}} 
\end{example}

\revision{In order to \doublerev{ensure that the encoding represents a unique tile, } 
along with the positions of 1s, we also 
\doublerev{need to identify columns and rows that are parts of a tile  
but contain only 0s. We do that by storing IDs of all items and transactions that are part of a tile (i.e., relevant to it) using the facts \texttt{relitem(id,i)} and \texttt{reltrans(id,t)} respectively.}}
\begin{example}
\revision{The facts in Example~\ref{ex:tileenc} and $\{$\texttt{relitem(1,1),relitem(1,2),relitem(1,3),}\\\texttt{relitem(1,4), reltrans(1,1), reltrans(1,2)}$\}$ uniquely represent the tile \doublerev{in the middle} 
of Figure~\ref{fig:tileenc}. Analogously, if we instead 
$\{$\texttt{relitem(1,1), relitem(1,2), reltrans(1,3)}$\}$ are added to the facts in Example~\ref{ex:tileenc}, then the encoding of the tile on the right-hand side of Figure~\ref{fig:tileenc} is obtained. \qed}
\end{example}

 \revision{Note that if there exists a $1$ in the original data that is not a part of any tile, no method will be able to cover it. Hence such $1$s 
are accumulated into a constant error, and \doublerev{we} ignore them in our setting.}

\begin{figure}
\begin{center}
\begin{tikzpicture}
        \matrix [matrix of math nodes,left delimiter=(,right delimiter=)] (m)
        {
            1 &0 & 0 &1  \\               
            0 &0 & 0 &0  \\               
            1 &0 & 0 &1  \\           
        };  
    \end{tikzpicture}
\begin{tikzpicture}
        \matrix [matrix of math nodes,left delimiter=(,right delimiter=)] (m)
        {
            1 &0 & 0 &1  \\               
            0 &0 & 0 &0  \\               
            1 &0 & 0 &1  \\           
        };  
        \draw[color=red] (m-1-1.north west) -- (m-1-4.north east) -- (m-1-4.south east) -- (m-1-1.south west) -- (m-1-1.north west);
        \draw[color=red] (m-3-1.north west) -- (m-3-4.north east) -- (m-3-4.south east) -- (m-3-1.south west) -- (m-3-1.north west);
    \end{tikzpicture}
\begin{tikzpicture}
        \matrix [matrix of math nodes,left delimiter=(,right delimiter=)] (m)
        {
            1 &0 & 0 &1  \\               
            0 &0 & 0 &0  \\               
            1 &0 & 0 &1  \\           
        };  
        \draw[color=red] (m-1-1.north west) -- (m-3-1.south west) -- (m-3-1.south east) -- (m-1-1.north east) -- (m-1-1.north west);
        \draw[color=red] (m-1-4.north west) -- (m-1-4.north east) -- (m-3-4.south east) -- (m-3-4.south west) -- (m-1-4.north west);
    \end{tikzpicture}
\end{center}
\caption{\revision{Examples of tiles in a matrix}}
\label{fig:tileenc}
\end{figure}

  In (1)--(3) of Listing~\ref{lst:tiles} we collect tiles' IDs and locations of ones using the dedicated predicates \texttt{tileid} and \texttt{one} respectively. Than in (4)--(5) for every tile we perform a guess on whether to include it into the tiling. In lines (6)--(9), we count the number of 1s that are outside of the tiling, while in lines (10)--(13), the number of 0s inside the tiling is computed. Finally, the total error is determined in lines (14)--(17) and its admissibility based on the given threshold is checked in lines (18)--(19).

This ASP program can be exploited to solve Boolean matrix factorization, when used together with existing candidate generation approaches, or to solve Boolean linear programming instances. In fact, in Boolean linear programming, the design matrix $\mA$ is given as a part of the problem instance, hence for that even a purely declarative programming solution will suffice. 


\section{Evaluation}\label{sec:eval}
In this section we evaluate the proposed hybrid approach by comparing it to the existing declarative pattern mining methods: ASP model for sequences \cite{DBLP:conf/ijcai/GebserGQ0S16} and Dominance Programming (DP) \cite{dp2013}. We do not consider the itemset mining ASP model \cite{DBLP:conf/lpnmr/Jarvisalo11},  
since 
it focuses only on frequent itemset mining and is not applicable to the 
construction of condensed representations under constraints as explained and addressed in \cite{dp2013}. Moreover, we do not perform comparison with dedicated algorithms designed for a specific problem type; these are known to be more efficient \revision{(in terms of runtime and memory usage)} than declarative mining approaches \cite{DBLP:conf/cpaior/NegrevergneG15}, yet obviously less flexible \revision{(in terms of modelling flexibility, i.e., how general the method is and how easily it can be modified to model a variaton of the problem)}. 

More specifically, we investigate the following experimental questions. 

\begin{itemize} \setlength{\itemsep}{5pt}
  \item \qone: How does the runtime of our method compare to the existing ASP-based sequence mining models?
  \item \qtwo: What is the runtime gap between the specialized mining languages such as dominance programming and our method?
  \item \qthree: What is the influence of local constraints on the 
runtime of 
our method? 
  \item \qfour: How does the choice of an ASP solver influence the overall performance of the system?
  \item \qfive: How does the hybrid system perform on highly structured graph mining task in comparison to other logic-based mining systems?
\item \qsix: How does the hybrid method perform on the approximate tiling problem?
\end{itemize}

For \qone we compare our work with the ASP-based model by 
\citeN{DBLP:conf/ijcai/GebserGQ0S16}. For \qtwo we measure 
the runtime difference between specialized itemset mining languages \cite{dp2013}
and our ASP-based model.  To address \qthree we estimate the runtime effect of adding local constraints. Moreover, for \qfour we compare the performance of our system with two state-of-the-art ASP solvers: Clasp and WASP. Regarding \qfive, we perform the comparison of our hybrid mining system against the existing ILP-based method for graph mining. Finally, to tackle \qsix as a proof of concept we test the effectiveness of our hybrid method for solving the problem from Definition~\ref{def:probtile}. 

We report our evaluation results on 2 transaction datasets\!\footnote{From \url{https://dtai.cs.kuleuven.be/CP4IM/datasets/}.}: \textit{Mushrooms} (8124 transactions/119 items) and \textit{Vote} (435/48), 3 sequence datasets (full)\!\footnote{From \url{https://dtai.cs.kuleuven.be/CP4IM/cpsm/datasets.html}.}: \textit{JMLR} (788 sequences/3847 distinct symbols), \textit{Unix Users} (9099/4093), and \textit{iPRG} (8628/21), 3 graph datasets\!\footnote{From \url{https://github.com/amaunz/ofsdata}}: \textit{Yoshida} (265 graphs/20 avg. vertices/23 avg. edges/9 distinct labels), \textit{Nctrer} (232/19/20/9) and \textit{Bloodbarr}(413/21/23/9), and 2 datasets for the tiling problem: \emph{Divorce}\footnote{\url{https://sparse.tamu.edu/Pajek/divorce}}  (50 transactions/9 items) and \emph{Glass}\footnote{\url{http://cgi.csc.liv.ac.uk/\textasciitilde frans/KDD/Software/LUCS-KDD-DN/DataSets/dataSets.html}} (214/48).
%
All experiments have been performed on a desktop with Ubuntu 14.04, 64-bit environment, Intel Core i5-3570 4xCPU 3.40GHz and 8GB memory using clingo 4.5.4 \!\footnote{\url{http://potassco.sourceforge.net}} C++14 for the itemset/sequence wrapper and python 2.7 for the graph wrapper. In the evaluation we used 
the latest available version 2 of WASP\footnote{\url{https://github.com/alviano/wasp}}. 
The timeout was set to one hour. Free pattern mining demonstrates the same runtime behavior as closed, due to the symmetric encoding, and is thus omitted. 
\newcommand{\scalefigures}{1}
\begin{figure}[tb]
  \centering
  \begin{subfigure}[t]{0.49\textwidth}
   \includegraphics[width=\scalefigures\textwidth]{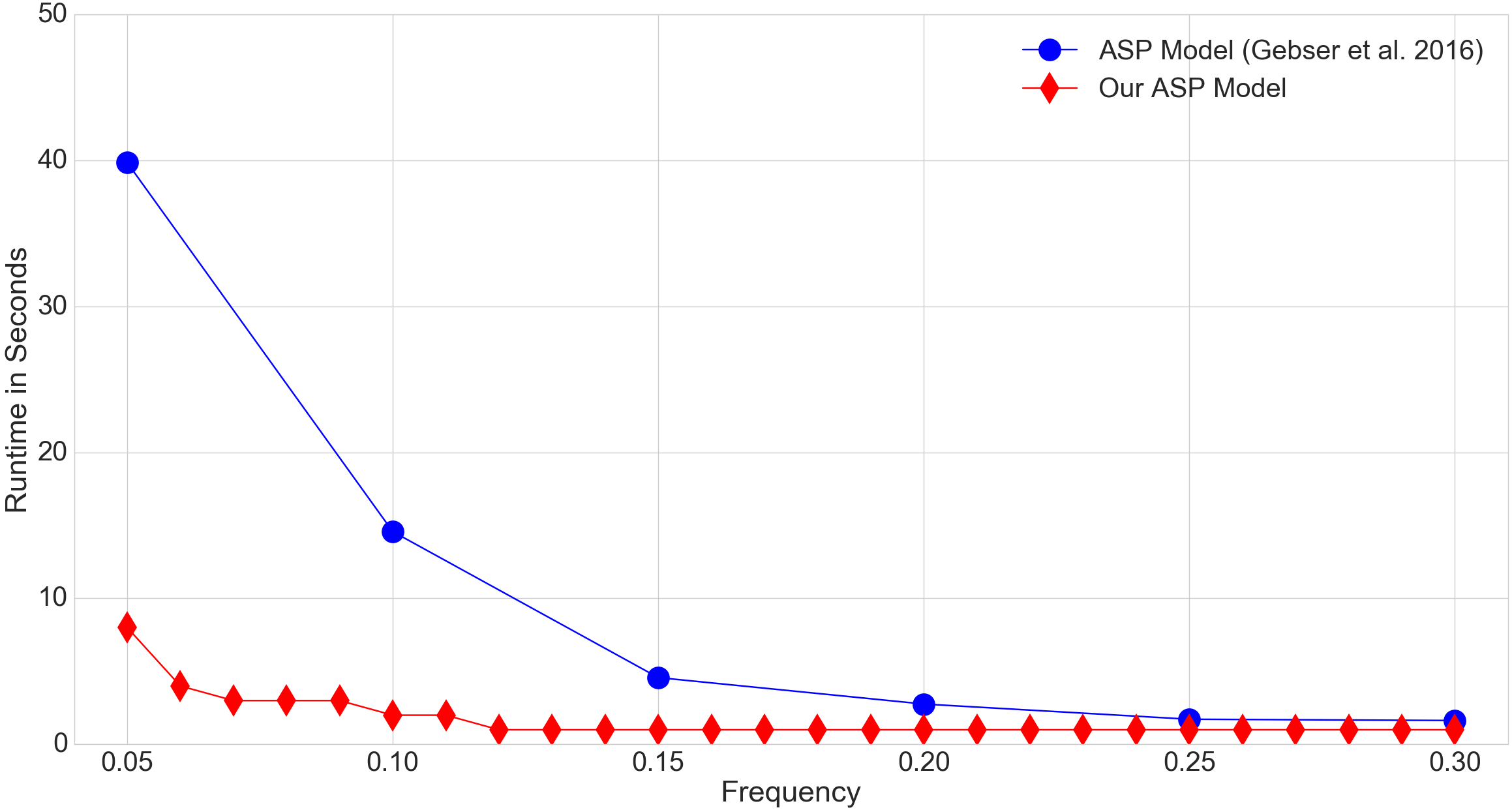}
   \caption{Comparing with ASP sequence model \cite{DBLP:conf/ijcai/GebserGQ0S16} on the 200 generated sequences (closed)}
    \label{fig:sequence_comparison}
  \end{subfigure}
 \hfill
  \begin{subfigure}[t]{0.49\textwidth}
   \includegraphics[width=\scalefigures\textwidth]{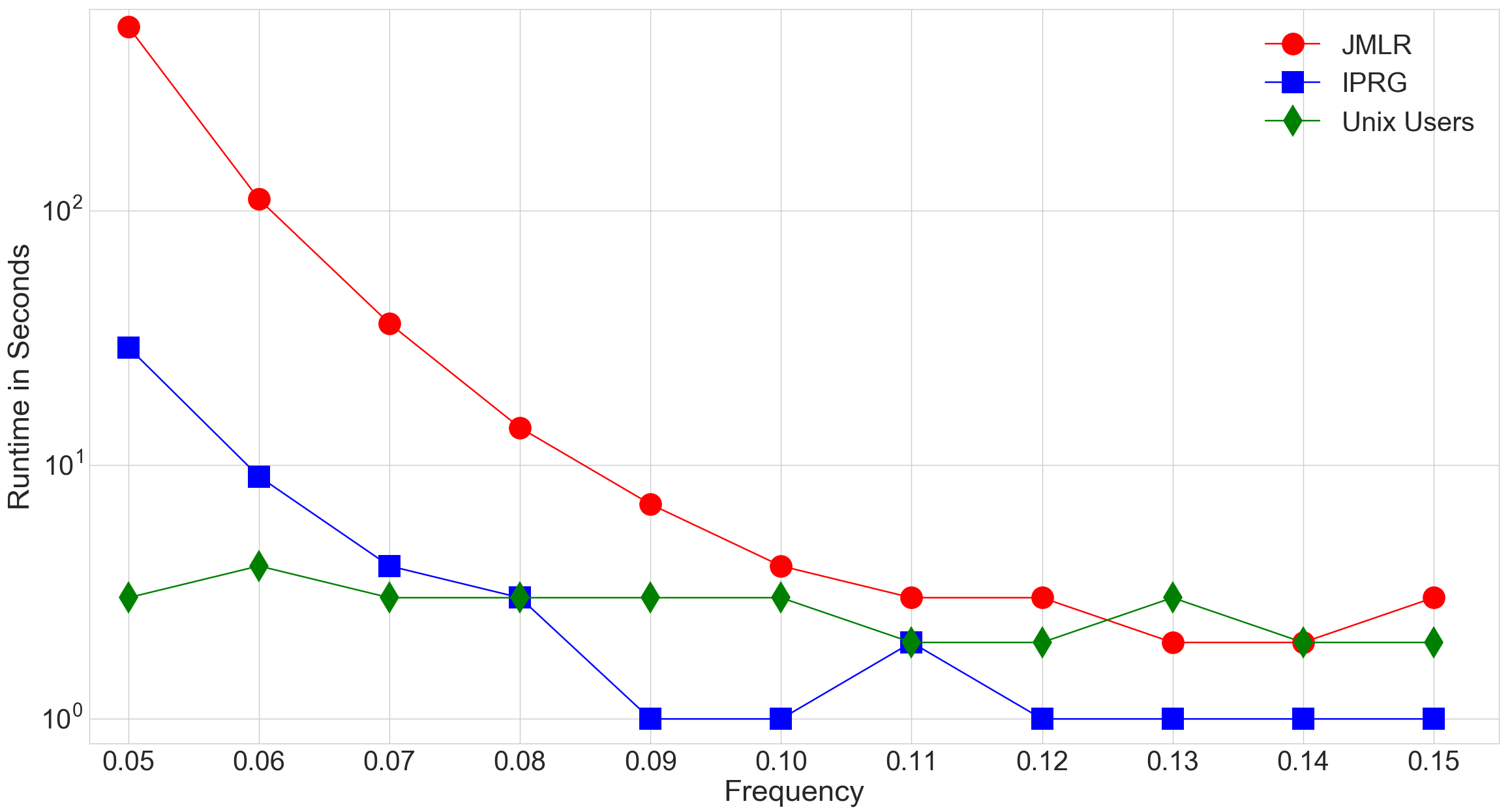}
   \caption{Maximal sequence patterns} 
    \label{fig:maximal}
  \end{subfigure}
 \hfill
  \begin{subfigure}[t]{0.49\textwidth}
   \includegraphics[width=\scalefigures\textwidth]{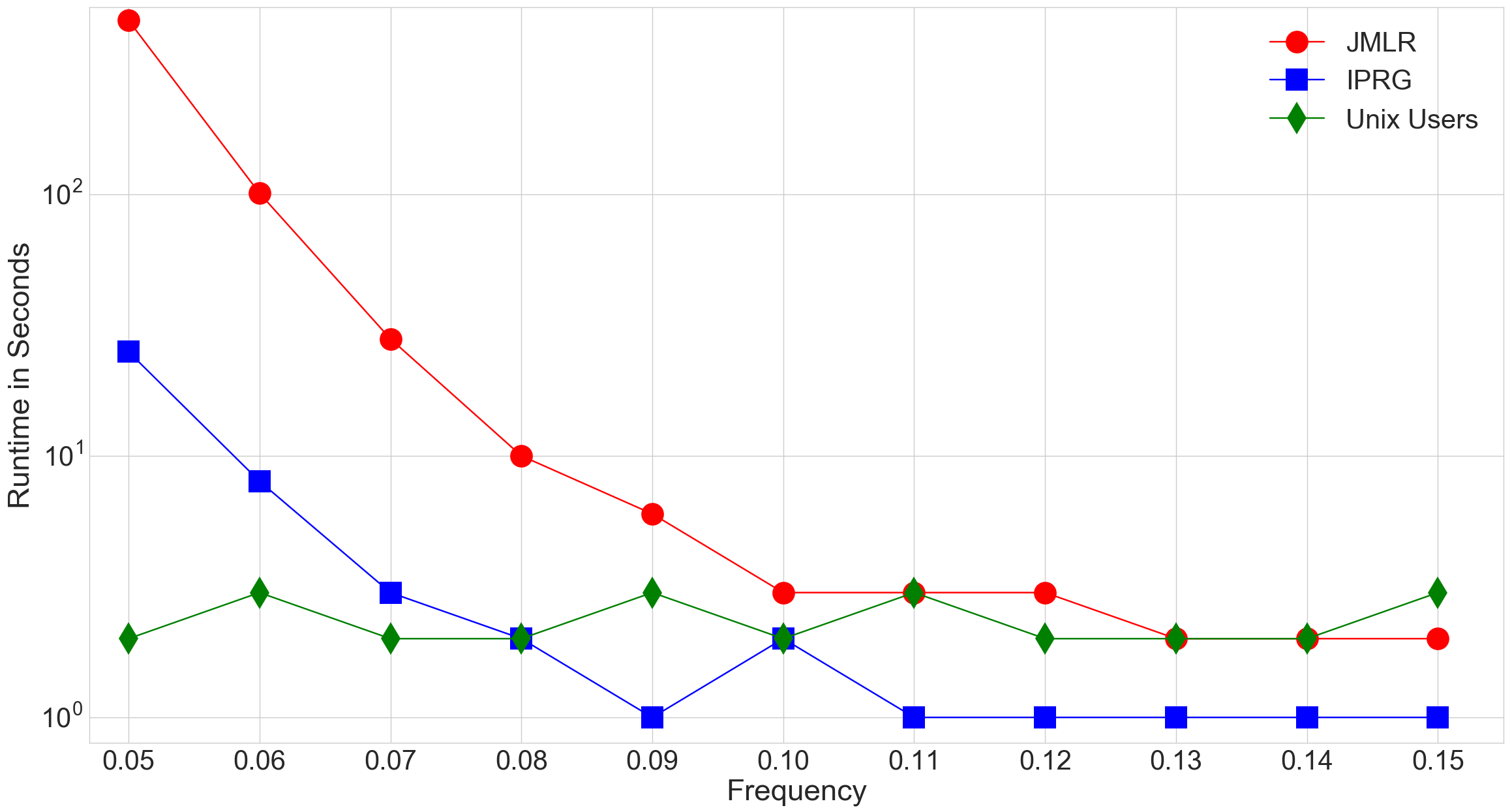}
   \caption{Closed sequence patterns} 
    \label{fig:closed}
  \end{subfigure}
\hfill
  \begin{subfigure}[t]{0.49\textwidth}
   \includegraphics[width=\scalefigures\textwidth]{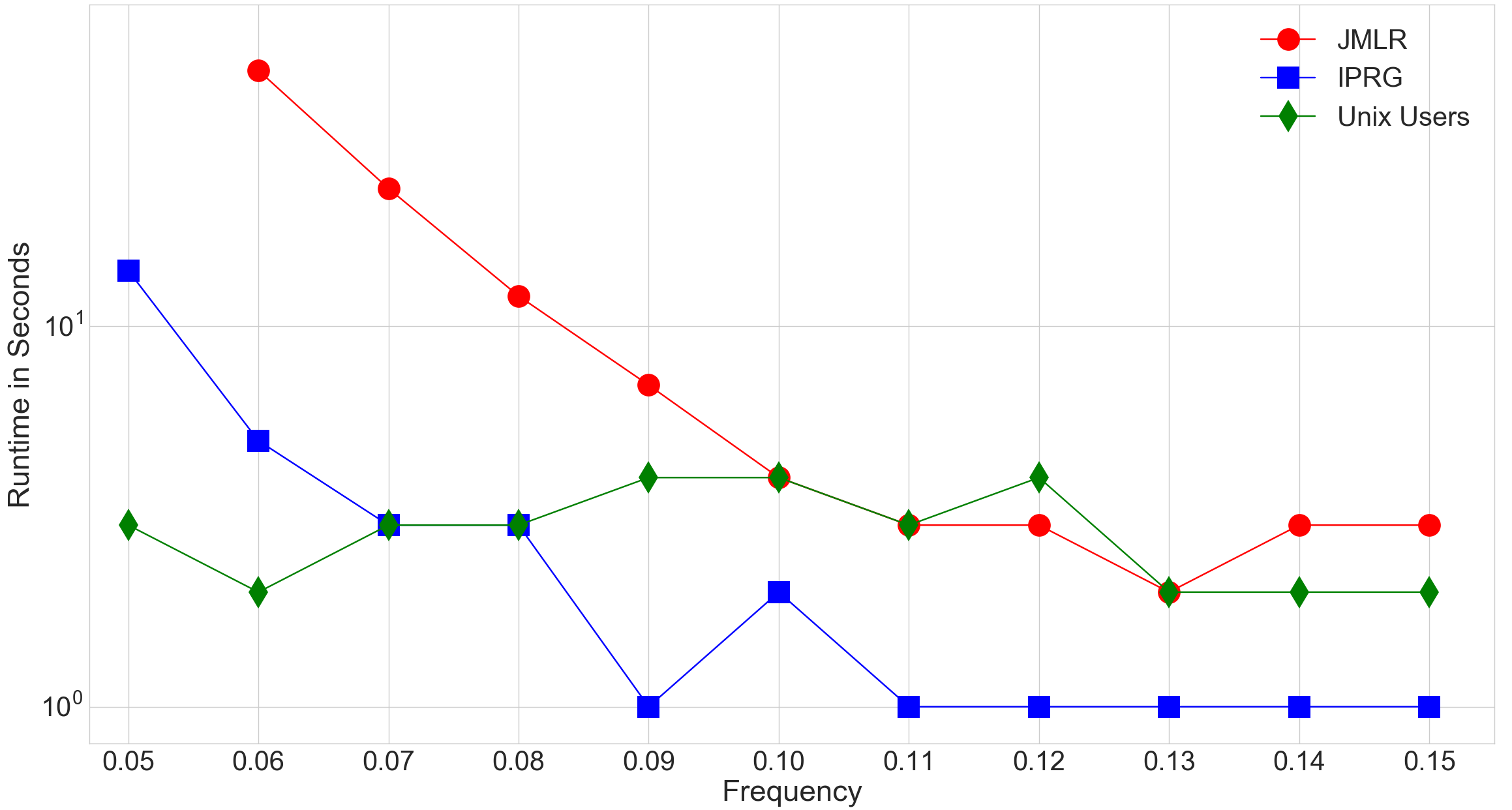}
   \caption{Skyline sequence patterns} 
    \label{fig:skyline}
  \end{subfigure}
  \caption{Investigating \qone: comparison with pure ASP model (\ref{fig:sequence_comparison}) and maximal (\ref{fig:maximal}),  closed (\ref{fig:closed}), and skyline (\ref{fig:skyline}) sequence mining on  JMLR, Unix Users, and iPRG datasets.}
  \label{fig:qone}
\end{figure}

To investigate \qone, in Figure~\ref{fig:sequence_comparison}, we compare the ASP model \cite{DBLP:conf/ijcai/GebserGQ0S16} with our method on the default 200 sequence sample, generated by the tool\footnote{\url{https://sites.google.com/site/aspseqmining}} of \citeN{DBLP:conf/ijcai/GebserGQ0S16}. We 
performed the comparison on the synthetic data, as the sequence-mining model \cite{DBLP:conf/ijcai/GebserGQ0S16} failed to compute condensed representations on any of the standard sequence datasets for any support threshold value within the timeout. One can observe that our method consistently outperforms the purely declarative approach of \citeN{DBLP:conf/ijcai/GebserGQ0S16} and the advantage naturally becomes more apparent for smaller frequency threshold values. 
 
In Figures~\ref{fig:maximal},~\ref{fig:closed} and~\ref{fig:skyline} (the point $0.05$ for JMLR is a timeout), we present the runtimes of our method for \emph{maximal, closed} and \emph{skyline} sequential pattern mining settings on JMRL, Unix Users and iPRG datasets. In contrast to \citeN{DBLP:conf/ijcai/GebserGQ0S16}, our method managed to produce results on all of these datasets for reasonable threshold values within a couple of minutes.

\begin{figure}[tb]
  \centering
  \begin{subfigure}[t]{0.49\textwidth}
   \includegraphics[width=\scalefigures\textwidth]{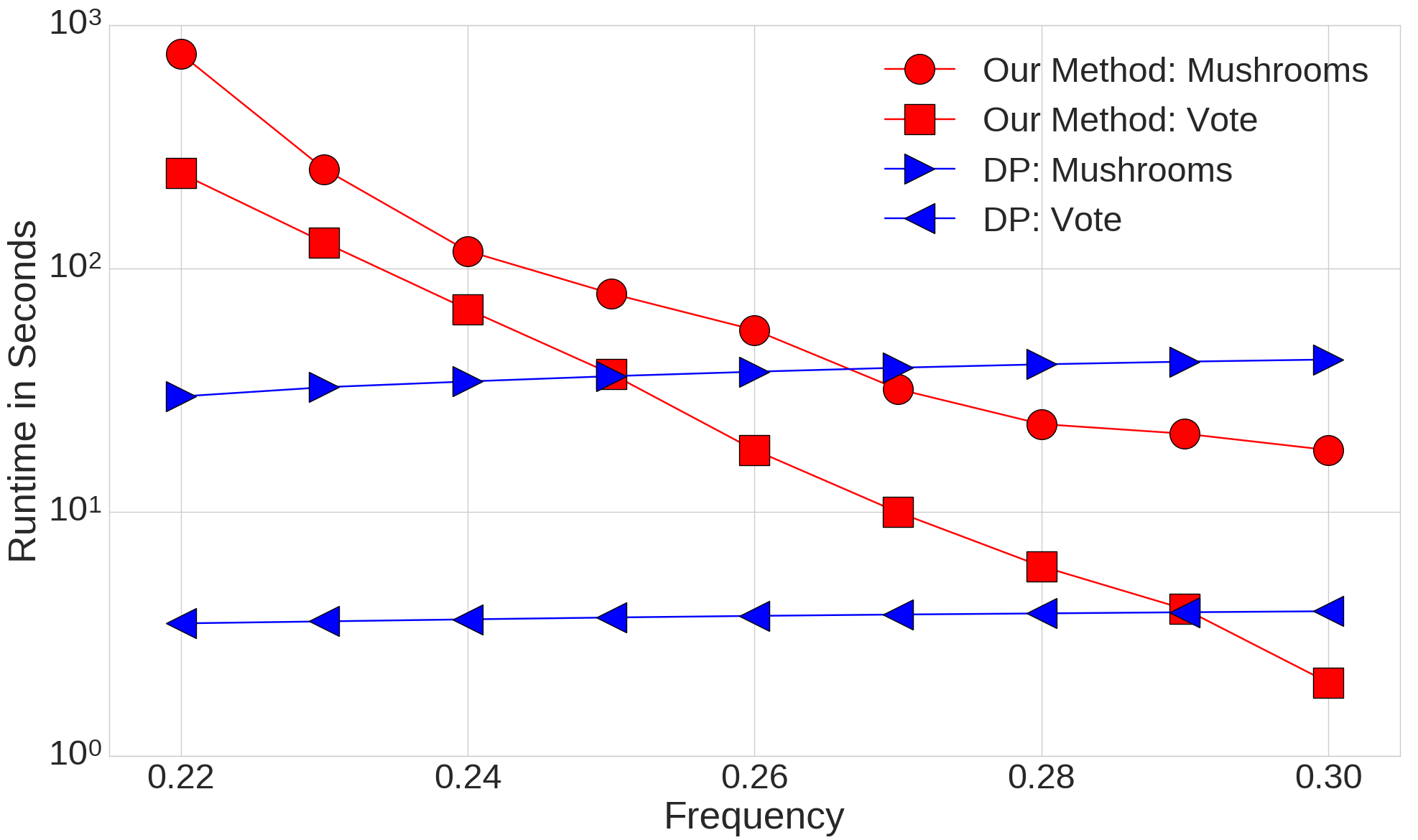}
   \caption{Maximal itemset mining: comparing with DP on Vote and Mushrooms}
    \label{fig:dp_maximal}
  \end{subfigure}
 \hfill
  \begin{subfigure}[t]{0.49\textwidth}
   \includegraphics[width=\scalefigures\textwidth]{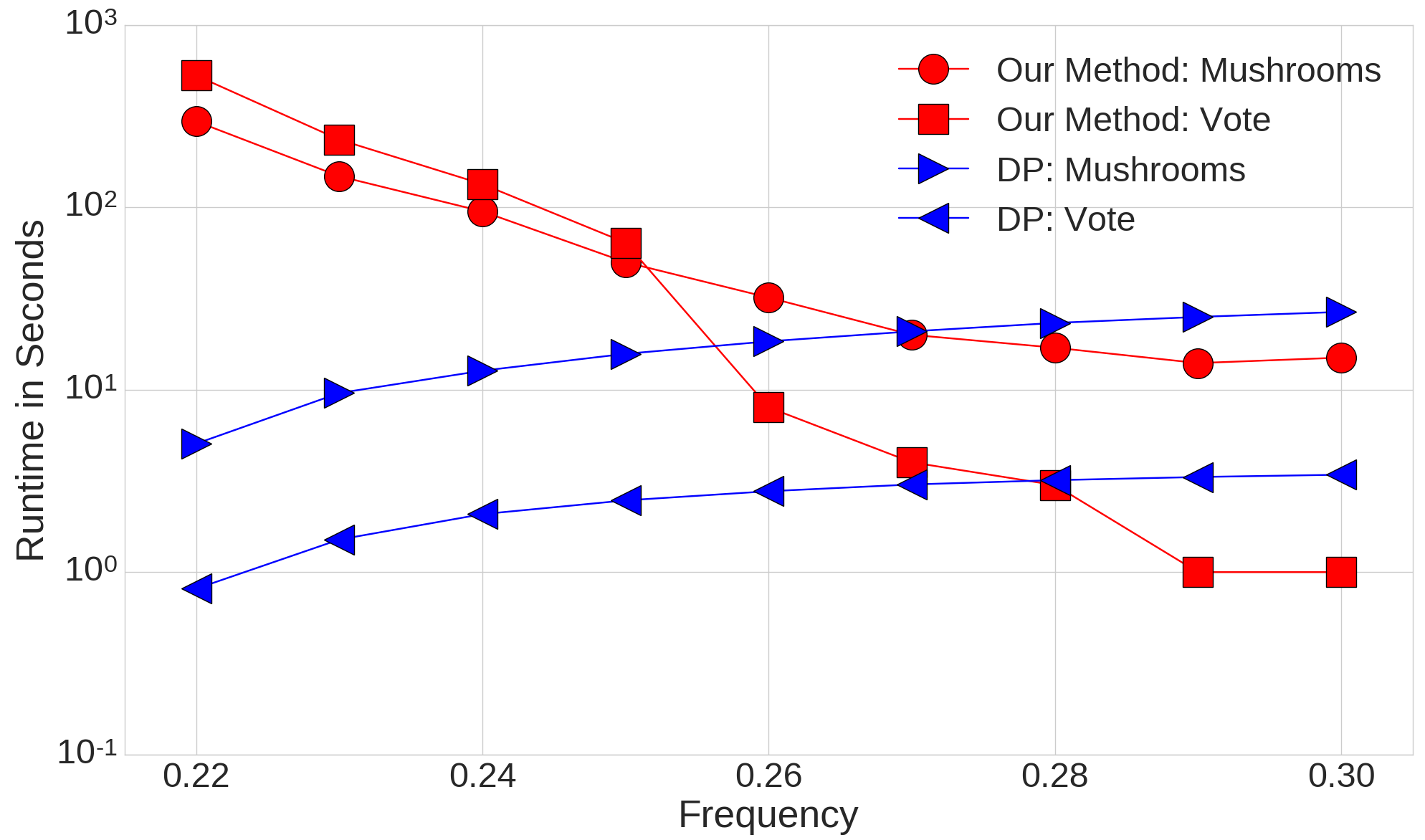}
   \caption{Closed itemset mining: comparing with DP on Vote and Mushrooms}
    \label{fig:dp_closed}
  \end{subfigure}
 \hfill
  \begin{subfigure}[t]{0.49\textwidth}
   \includegraphics[width=\scalefigures\textwidth]{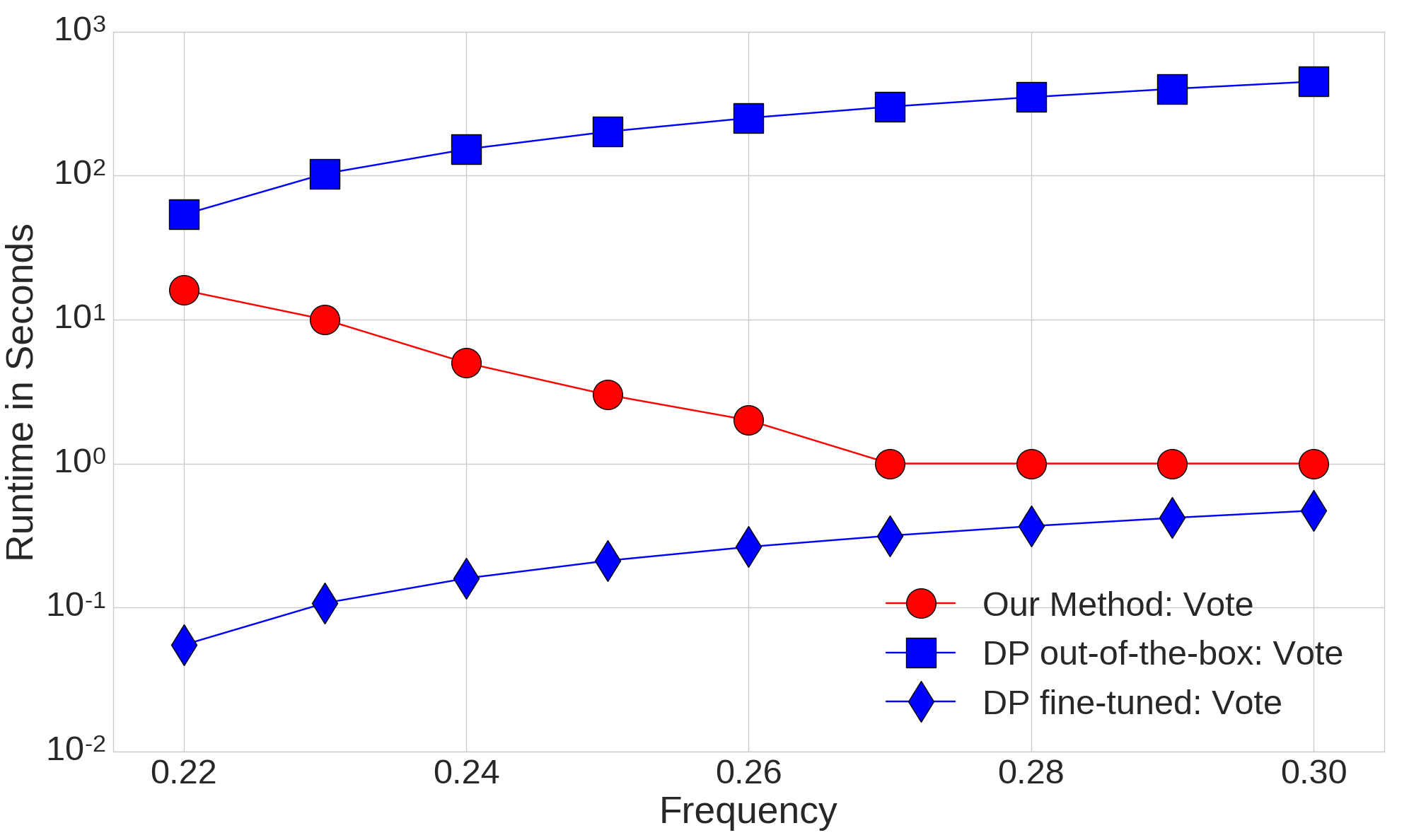}
   \caption{Skyline itemset mining: comparing with out-of-the-box and fine-tuned DP on Vote}
    \label{fig:dp_skyline}
  \end{subfigure}
\hfill
  \begin{subfigure}[t]{0.49\textwidth}
   \includegraphics[width=\scalefigures\textwidth]{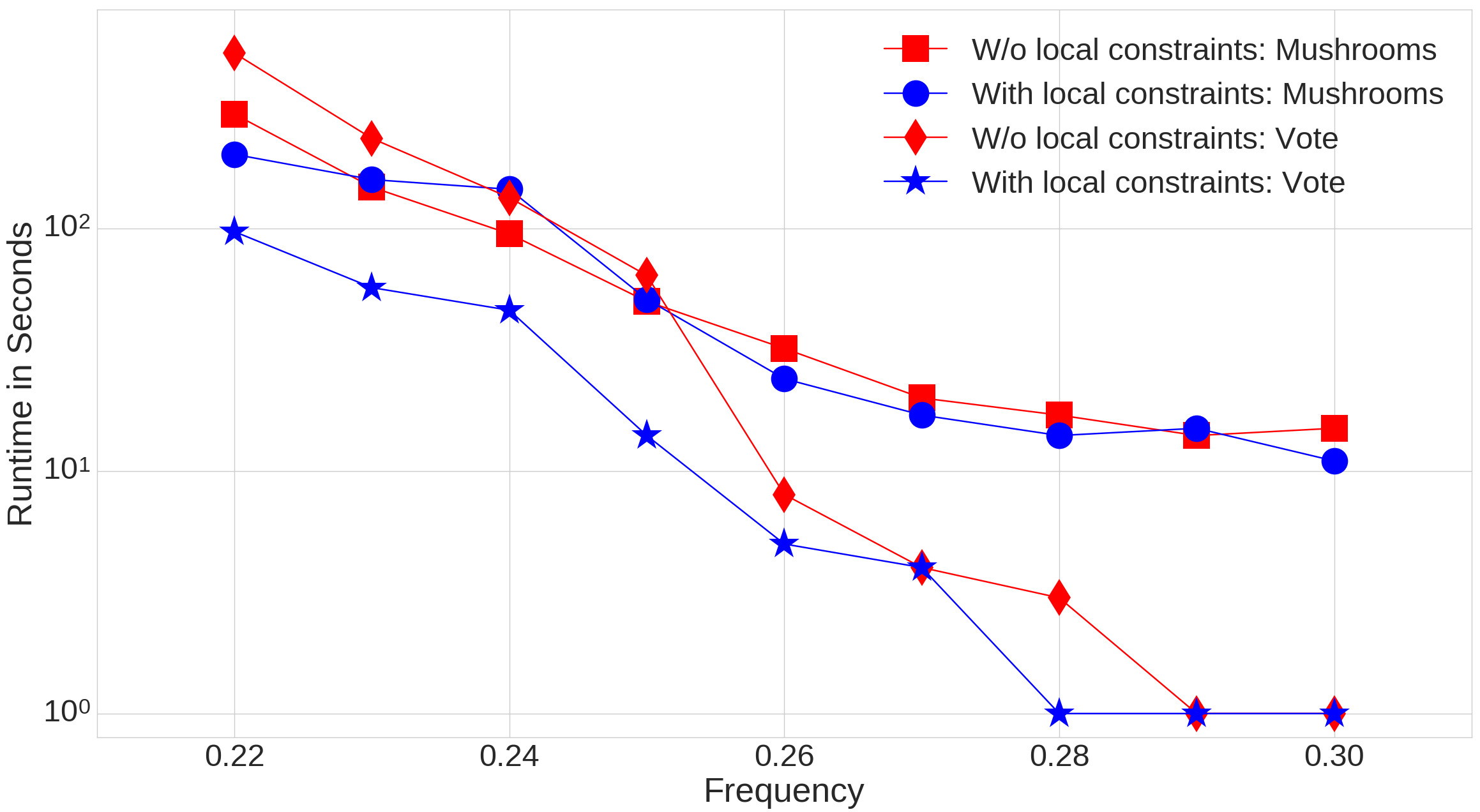}
   \caption{Closed itemset mining: our method with (w/o) local constraints on Vote and Mushrooms}
    \label{fig:local_constraints}
  \end{subfigure}
  \label{fig:qtwo_three}
\end{figure}

To investigate \qtwo, we compare out-of-the-box performance of DP \cite{dp2013} with our approach on maximal, closed and skyline itemset mining problems using standard datasets Vote and Mushrooms. As we see in Figures.~\ref{fig:dp_maximal} and \ref{fig:dp_closed}, on average, DP is one-to-two orders of magnitude faster; this gap is, however, diminishing as the minimum frequency increases. 
Surprisingly, our approach is significantly faster than DP out-of-the-box for skyline patterns (Figure~\ref{fig:dp_skyline}); this holds also for the Mushrooms dataset, not presented here. 

Fine-tuning parameters of DP by changing the order in which operators are applied within the system (skyline+ option) allowed to close this gap. With this adaptation DP demonstrates one-to-two orders of magnitude better performance, as can be seen in Figure~\ref{fig:dp_skyline}. However, fine-tuning such a system requires the understanding of its inner mechanisms or exhaustive application of all available options. 

To 
address \qthree we introduced three simple local constraints 
for the itemset mining setting from \qtwo: 
two size constraints $\textit{size(I)} > 2$ and $\textit{size(I)} < 7$ and a cost constraint: each item gets weight equal to its value with the maximal budget of $n$, which is set to $20$ in the experiments. 

In Figure~\ref{fig:local_constraints}, we present the results for closed itemset mining with and without local constraints (
experiments with other global constraints demonstrate a similar runtime pattern). 
Local constraints ensure 
better propagation and speed up the search. One of the key design features of our encoding is the filtering technique used to select candidate patterns among only valid patterns. Its effect can be clearly seen, e.g., for the Vote dataset in Figure~\ref{fig:local_constraints}, where for certain frequencies the runtime gap is close to an order of magnitude.

\begin{figure}[tb]
  \centering
  \begin{subfigure}[t]{0.49\textwidth}
   \includegraphics[width=\scalefigures\textwidth]{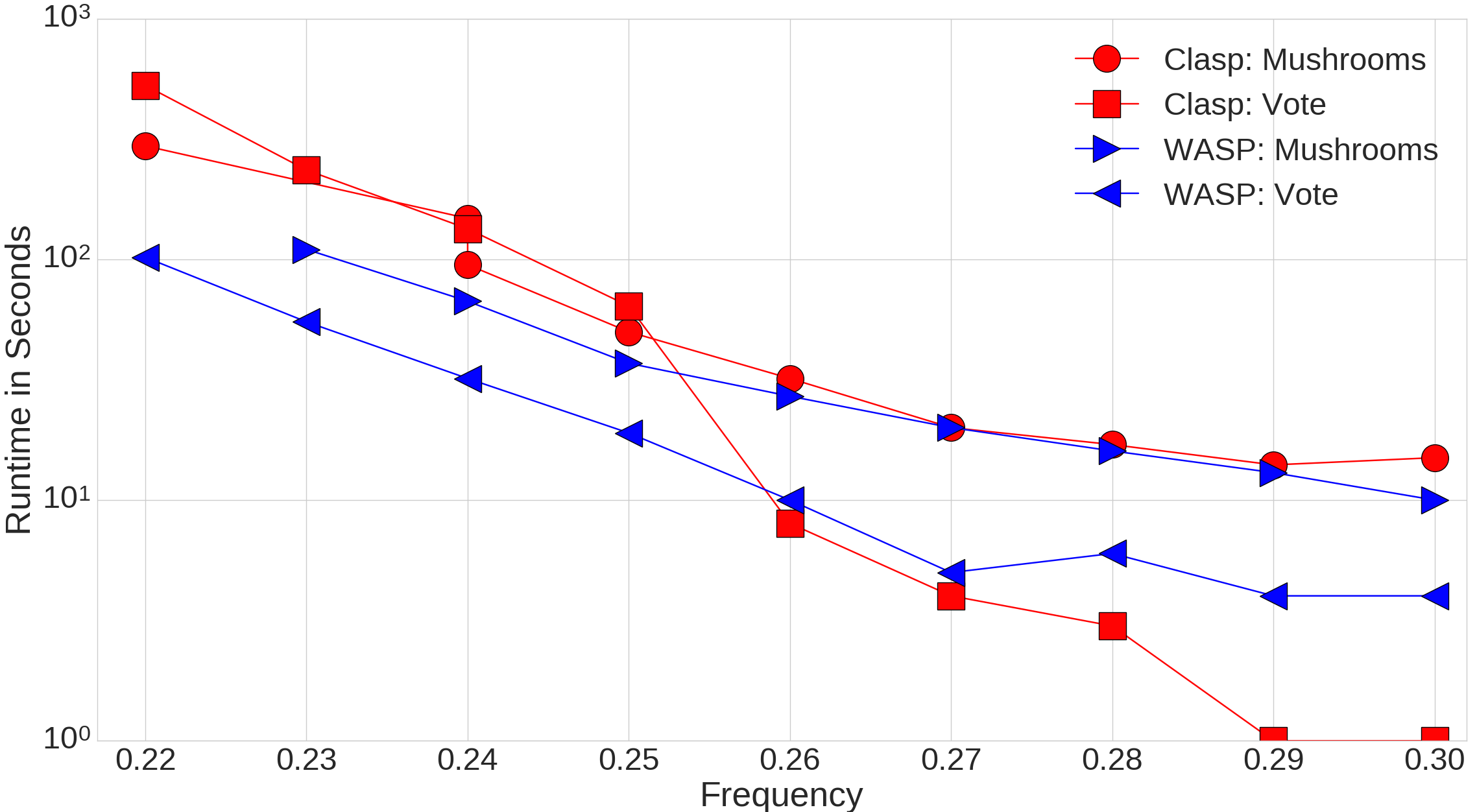}
   \caption{Itemset mining: closed patterns}
    \label{fig:itemset_wasp_closed_comparison}
  \end{subfigure}
 \hfill
  \begin{subfigure}[t]{0.49\textwidth}
   \includegraphics[width=\scalefigures\textwidth]{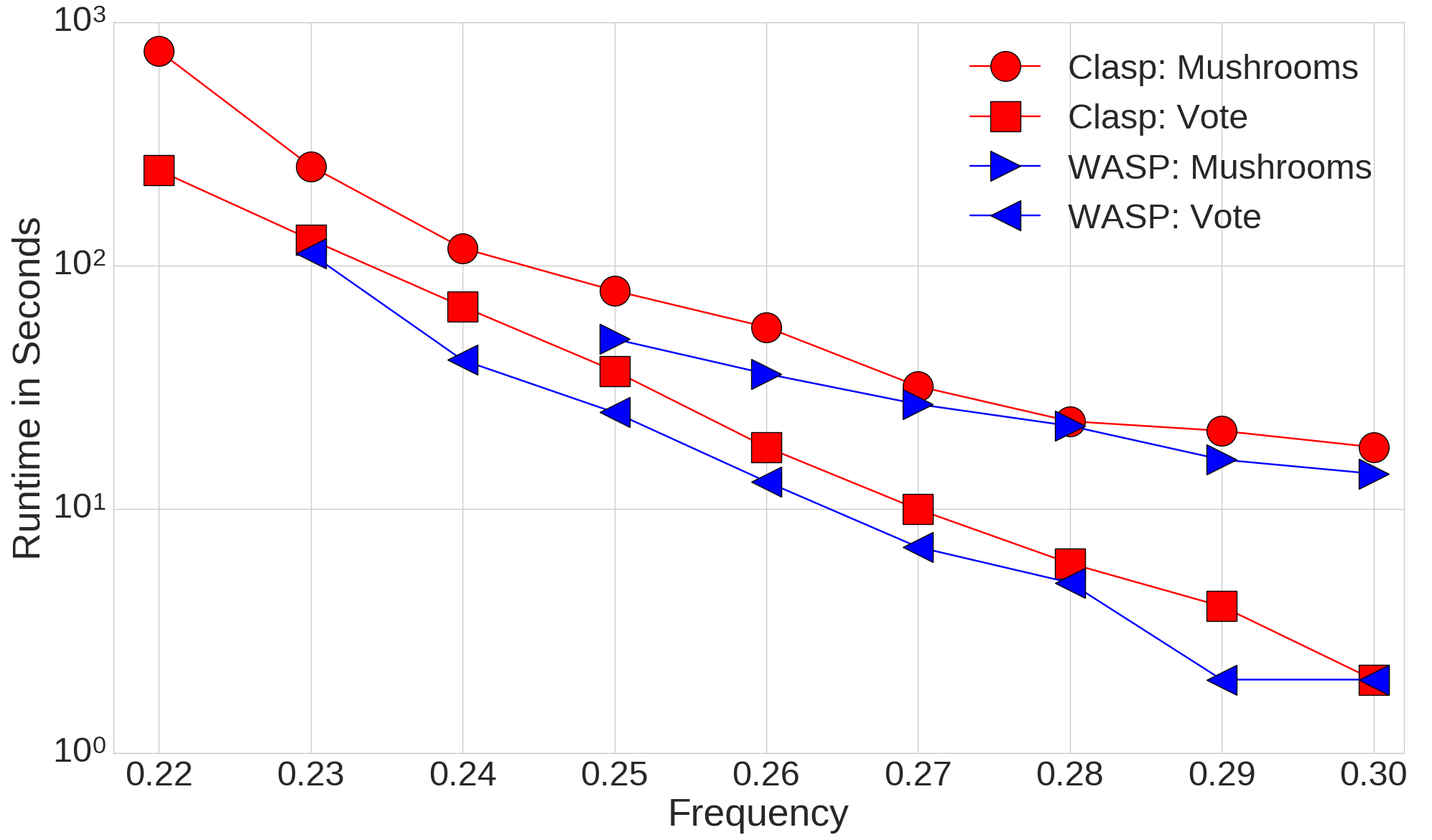}
   \caption{Itemset mining: maximal patterns} 
   \label{fig:itemset_wasp_max_comparison}
  \end{subfigure}
 \hfill
  \begin{subfigure}[t]{0.49\textwidth}
   \includegraphics[width=\scalefigures\textwidth]{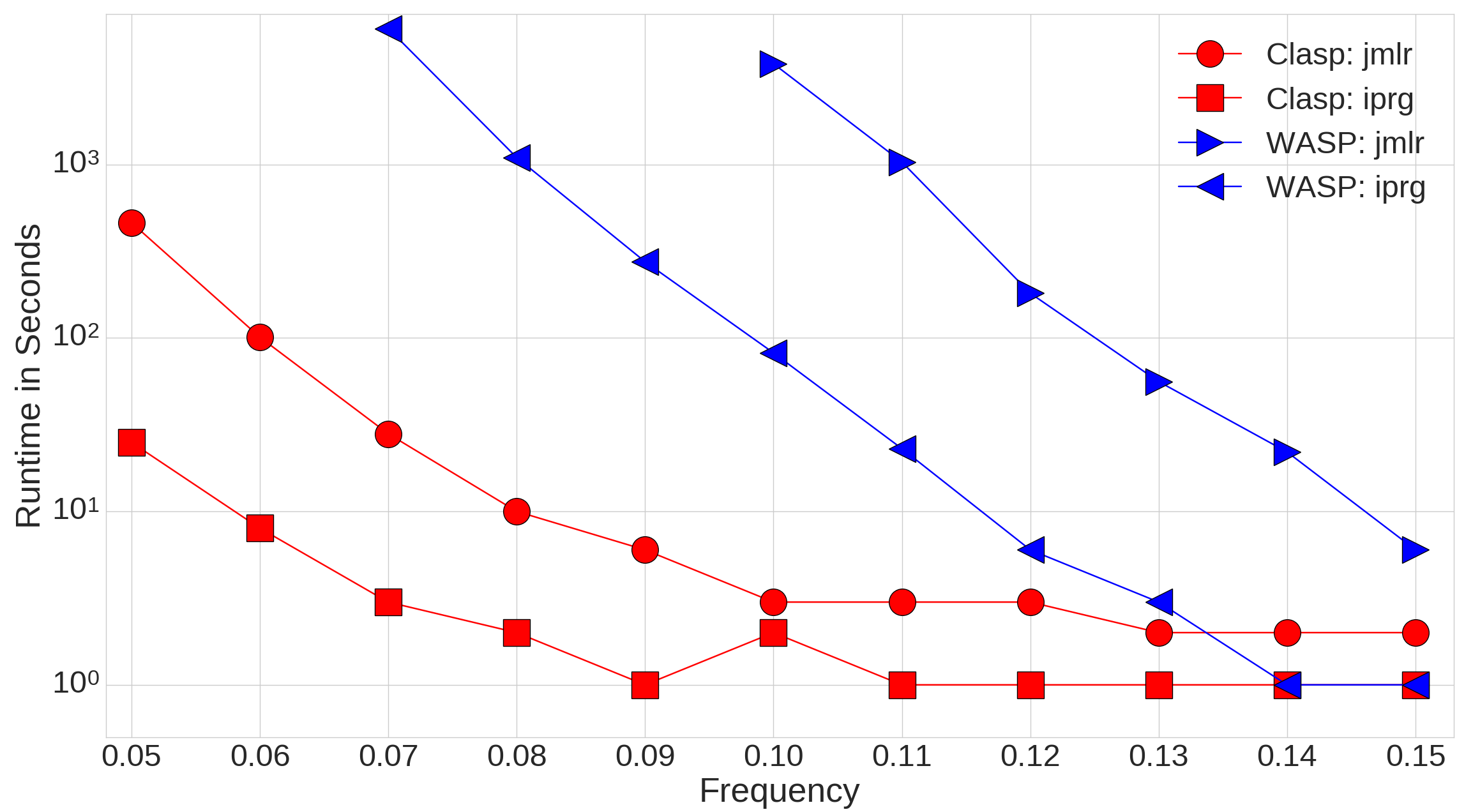}
	\caption{Sequence mining: closed patterns}
	\label{fig:seq_wasp_closed_comparison}
  \end{subfigure}
\hfill
  \begin{subfigure}[t]{0.49\textwidth}
   \includegraphics[width=\scalefigures\textwidth]{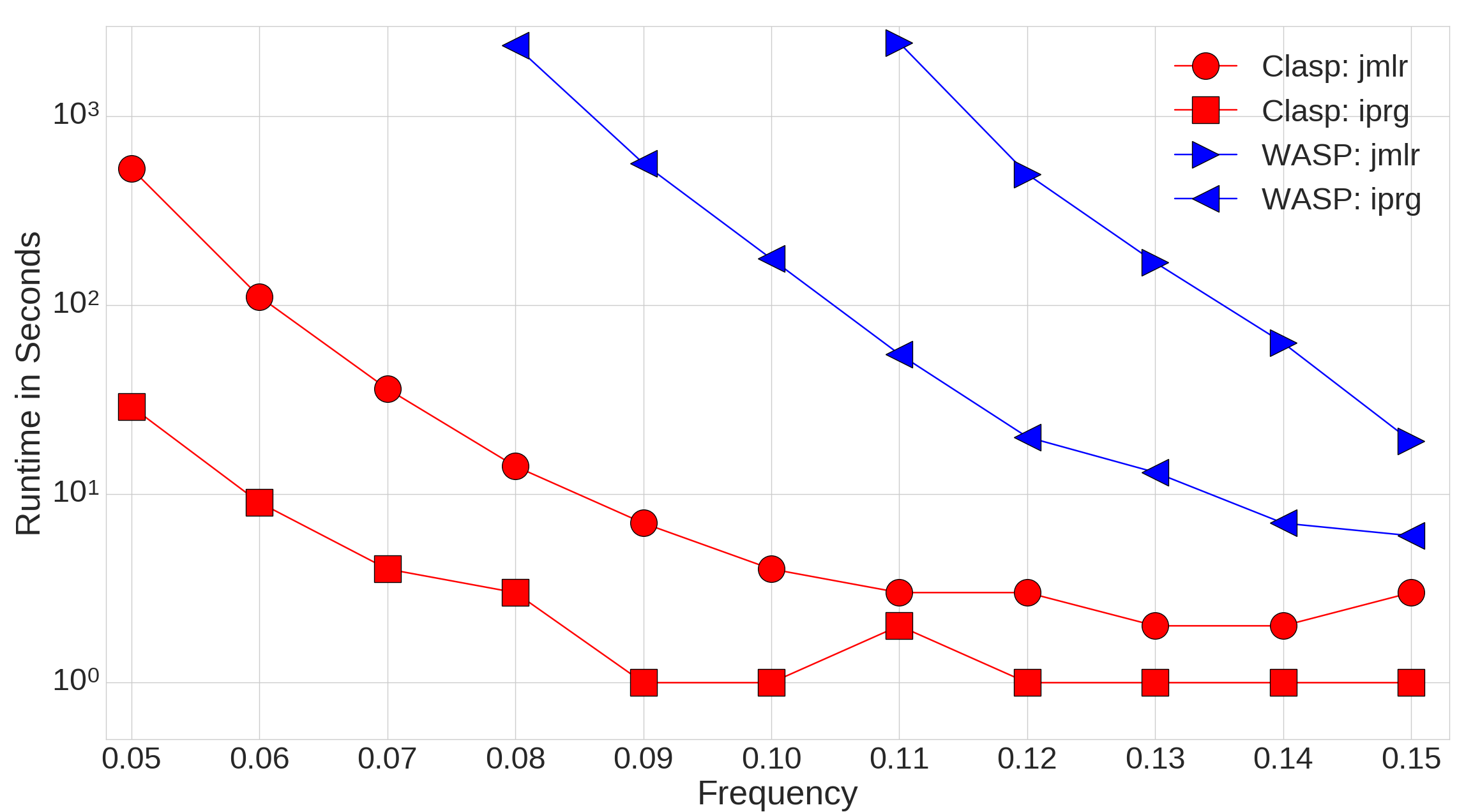}
   \caption{Sequence mining: maximal patterns}
    \label{fig:seq_wasp_max_comparison}
  \end{subfigure}
    \label{fig:wasp_comparison}
	\caption{\qfour: comparing ASP solvers performance -- WASP and Clasp on itemset and sequence mining tasks}
\end{figure}

To analyze \qfour, we have replaced Clasp solver with WASP \cite{wasp} in our system. As we see in Figure~\ref{fig:itemset_wasp_closed_comparison}, for mining closed itemsets two systems perform on par except for a timeout of WASP on mushrooms with frequency 0.22. However, we already can observe significant difference in performance on maximal patterns for both itemsets and sequences. A similar behavior can be seen in the setting of closed sequences mining: there is at least an order of magnitude difference in performance. The runtime gap cannot be explained by differences in grounding, since for both tasks Gringo has been used. However, we have noticed the difference in memory management: Clasp seems to be able to reason and keep track of significantly larger sets of patterns, while being economic and provident with memory. Since the task naturally allows one to generate problem instances of practically any size, we presume that our setting might be well-suited as one of the tests for ASP solvers' performance.

\begin{figure}[tb]
  \centering
  \begin{subfigure}[t]{0.49\textwidth}
   \includegraphics[width=\scalefigures\textwidth]{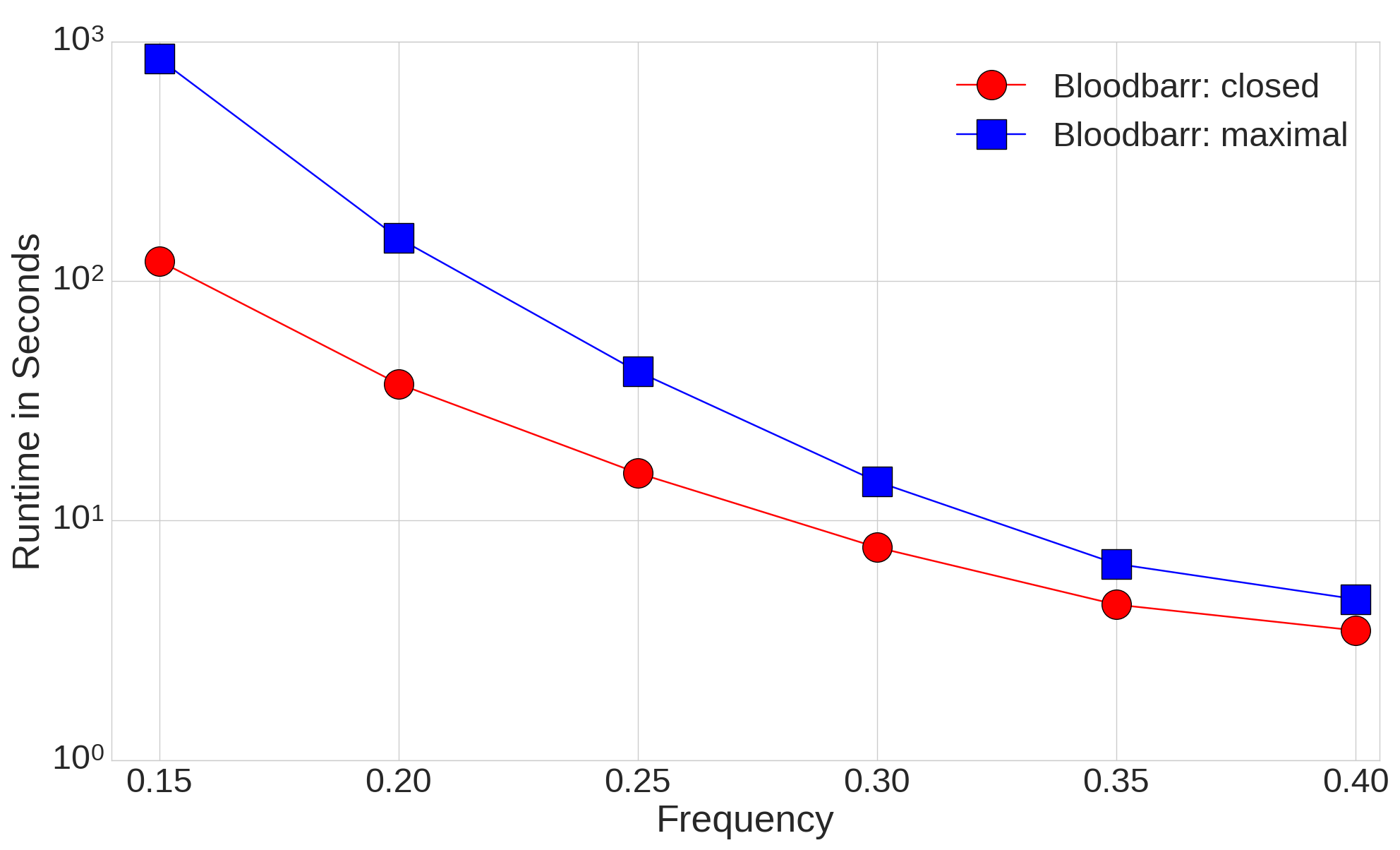}
   \caption{Bloodbarr dataset: maximal and closed graph patterns}
    \label{fig:bloodbarr}
  \end{subfigure}
 \hfill
  \begin{subfigure}[t]{0.49\textwidth}
   \includegraphics[width=\scalefigures\textwidth]{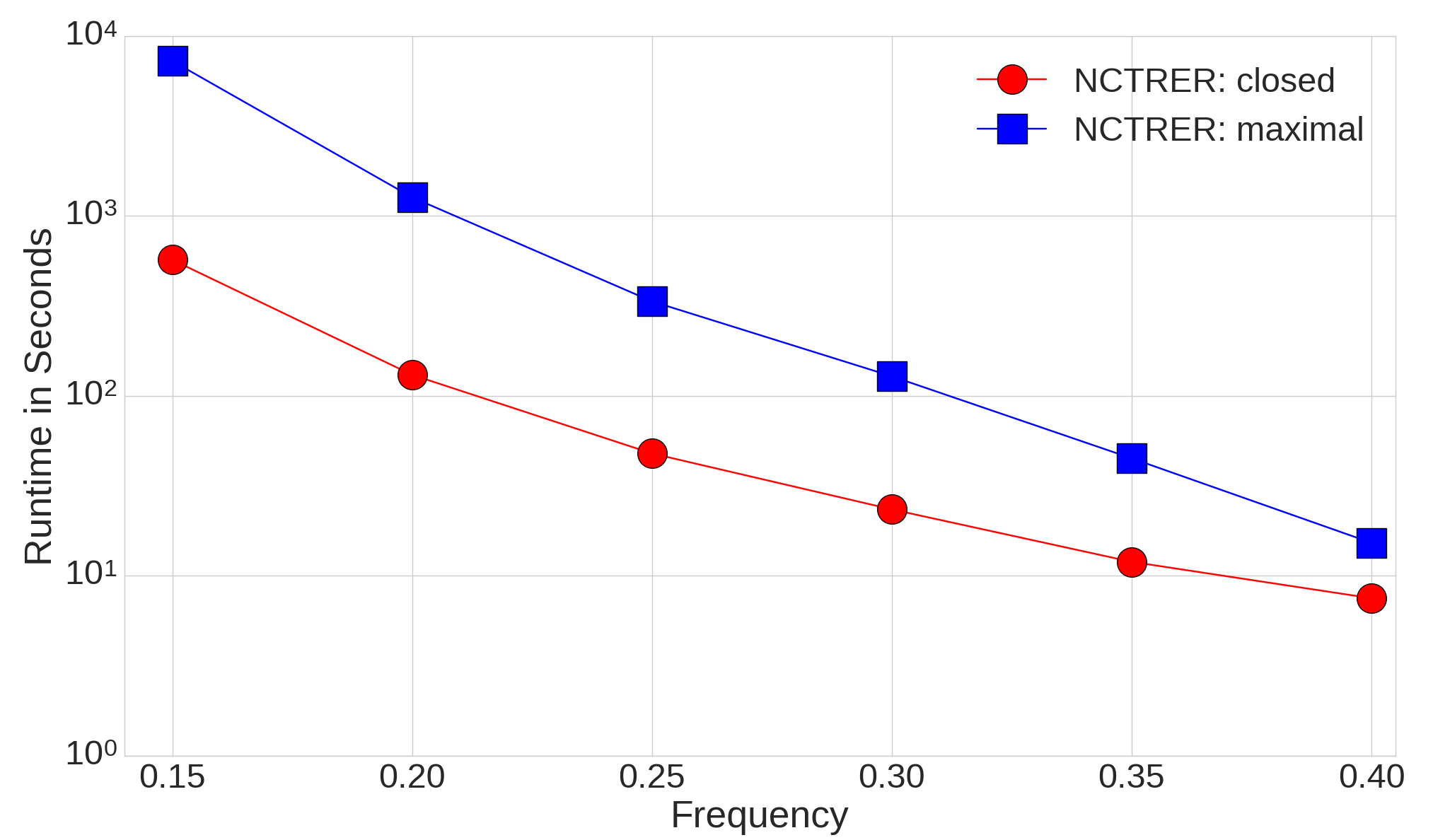}
   \caption{Nctrer dataset: maximal and closed graph patterns}
    \label{fig:nctrer}
  \end{subfigure}
 \hfill
  \begin{subfigure}[t]{0.49\textwidth}
   \includegraphics[width=\scalefigures\textwidth]{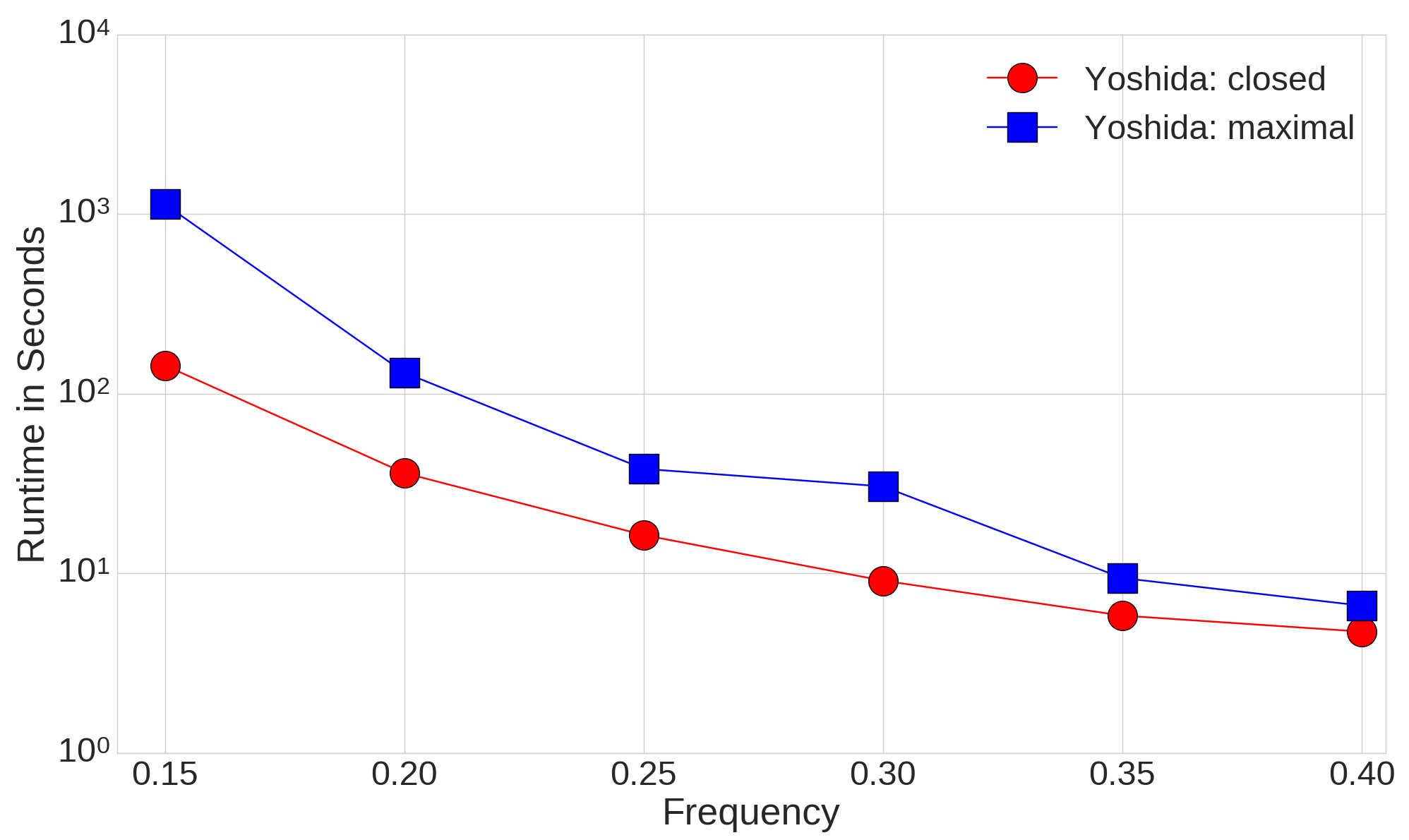}
   \caption{Yoshida dataset: maximal and closed graph patterns}
    \label{fig:yoshida}
  \end{subfigure}
\hfill
  \begin{subfigure}[t]{0.49\textwidth}
   \includegraphics[width=\scalefigures\textwidth]{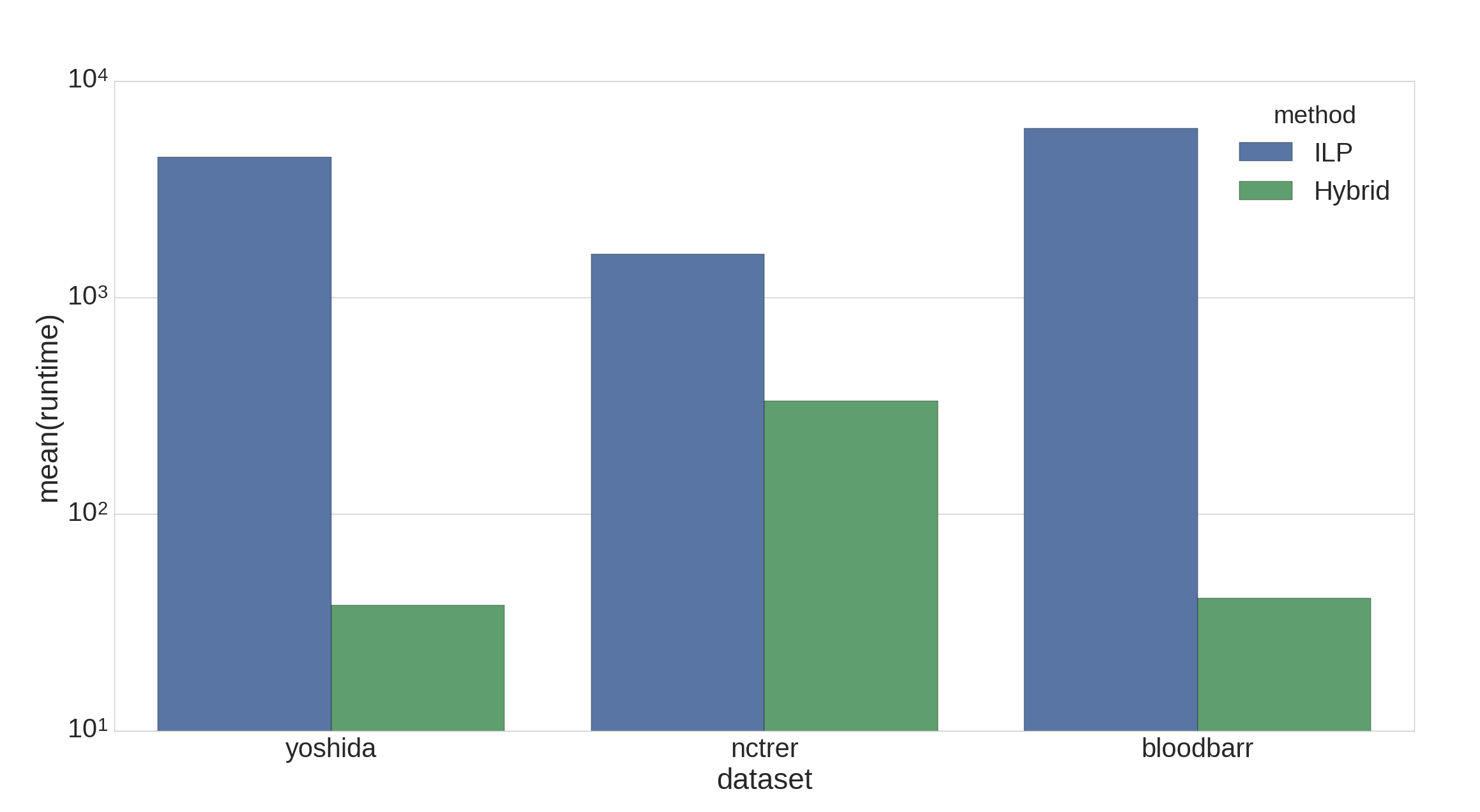}
   \caption{Comparison with logic programming ILP framework \cite{query_mining_ilp}. For the ILP framework time to find \textit{a maximal solution} is indicated, while for our system  time to enumerate all maximal graph patterns (frequency $0.2$) is shown.}
   \label{fig:ilp_comparison}
  \end{subfigure}
  \label{fig:qfive}
  \caption{Investigating \qfive: evaluation on standard graph datasets (a,b,c) and comparison to the declarative ILP system for graph mining (d). Log-scale on all plots}
\end{figure}

To address \qfive, we consider the standard graph mining datasets 
in Figs. \ref{fig:bloodbarr}, \ref{fig:nctrer} and \ref{fig:yoshida}. One can observe, that our system is indeed able to handle real-sized datasets and process hundreds 
or even thousands of graph patterns within the time-limit. Furthermore, the comparison with the purely declarative framework presented in Figure~\ref{fig:ilp_comparison} 
demonstrates the two orders of magnitude speedup in runtime. In addition, note that our system is able to enumerate all condensed graph patterns, while the purely declarative approach is not capable of doing that within the timeout.

To investigate \qsix, we tested our hybrid approach on the approximate tile selection problem from Def.~\ref{def:probtile}, exploiting Clasp reasoner for the ASP part of the algorithm. Since the considered problem is very computationally demanding, we did not set any timeout in this experimental setting. In Table~\ref{til:div} and Table~\ref{til:glass} we report the results for the Divorce and Glass datasets respectively. We compute the candidate tiles using an approach based on association confidences (see \citeNPS{miettinen08discrete}). Given the set of candidate tiles, we use our ASP encoding from Listing~\ref{lst:tiles} to find all tilings whose overall error is below a specified threshold. We compare this approach to a greedy method for finding \emph{any} tiling that has error below the threshold. The greedy tiling approach\revision{, proposed by~\citeN{miettinen08discrete},} will add the candidate tiles one-by-one until it has either found an admissible tiling, or it has exhausted all tiles. Note that, since the problem of finding the tiling is \NP-hard, the greedy method is not guaranteed to find an admissible tiling even if one exists. 

The first column in Table~\ref{til:div} (respectively Table~\ref{til:glass}) reports the number $n$ of candidate tiles and the error threshold value $\sigma$. In the second column we present the details of the solution found by the \revision{greedy} method, i.e., the number $k$ of tiles in the solution tiling and its overall error. The computation of the tilings by the greedy algorithm takes less than a second. 
In the third column, the results of our hybrid approach are provided. More specifically, we present the number $k$ of tiles and the overall error for the first solution found by the solver together with the running time, as well as the details of the optimal tiling. To compute the optimal tiling we enforce the solver to find all models of the given ASP program (their total number is likewise reported). 

 The candidate tiles are computed by the dedicated algorithm within a fraction of a second, and the major computational efforts are done by the ASP solver to find the final tiling. Moreover, observe that since the grounding step is the most time-consuming, the time difference between the first found solution and all solutions including the optimal one is actually neglectable.
Note that comparing the running time of the \revision{greedy} and our hybrid algorithms is not entirely fair, as the former is not complete (i.e., it might not find any tiling, satisfying the conditions imposed by the error threshold even if one exists), which is in contrast to our hybrid approach.

The \revision{greedy} algorithm is capable of computing the optimal tiling only for small instances of the Divorce dataset. Starting from $n=5$, our hybrid method outperforms the \revision{greedy} one with respect to the quality of the found solutions. For the Glass dataset the benefit of our method is apparent even for smaller instances.

\leanparagraph{Summary} In all experiments, Step 1 of our method contributes to less than 5\% of runtime. 
Overall, our approach can handle real world datasets for sequential pattern mining as demonstrated in \qone. In many cases 
its performance is 
close to the specialized mining languages, as 
shown in \qtwo. As demonstrated in \qthree various local constraints can be effectively incorporated into our encoding
bringing additional performance benefits. The choice of an ASP solver plays a crucial role in the overall performance of the system, as discussed in \qfour. In \qfive, it has been established that our approach leads to a significant speed up (of orders of magnitude) for the mining tasks with a complex structure, such as graph mining. Finally, the results of \qsix have proved the applicability of our hybrid framework for other computationally intensive data mining tasks, e.g., approximate tile selection problem, where our method is able to find solutions of higher quality.

\begin{table}[]
\centering
\begin{tabular}{rrrrrrrrrr}
\hline
\hline
\multicolumn{1}{l}{\multirow{3}{*}{$n$/$\sigma$}} & \multicolumn{2}{c}{\revision{greedy}}                                                           & \multicolumn{7}{c}{hybrid}                                                                                                                                                                                                                                                              \\ \cline{2-10} 
\multicolumn{1}{l}{}                         & \multicolumn{1}{c}{\multirow{2}{*}{$k$}} & \multicolumn{1}{c}{\multirow{2}{*}{error}} & \multicolumn{3}{c}{first}                                                      & \multicolumn{1}{c}{\multirow{2}{*}{$k$ opt}} & \multicolumn{1}{c}{\multirow{2}{*}{error opt}} & \multicolumn{1}{c}{\multirow{2}{*}{time all}} & \multicolumn{1}{l}{\multirow{2}{*}{$|$all solutions$|$}} \\ \cline{4-6}
\multicolumn{1}{l}{}                         & \multicolumn{1}{c}{}                   & \multicolumn{1}{c}{}                       & \multicolumn{1}{l}{$k$} & \multicolumn{1}{l}{error} & \multicolumn{1}{l}{time} & \multicolumn{1}{c}{}                       & \multicolumn{1}{c}{}                           & \multicolumn{1}{c}{}                          & \multicolumn{1}{l}{}                                 \\ \hline
3/59                                           & 1                                       & 51                                          & 1                      & 59                         & ~~38.961                    & 1                                           & 51                                              & ~~38.962                                         & 5                                                     \\ \hline
4/54                                           & 1                                       & 51                                          & 2                      & 51                         & ~~38.127                    & 1                                           & 51                                              & ~~38.135                                         & 8                                                     \\ \hline
5/60                                           & 2                                       & 51                                          & 2                      & 58                         & ~~50.616                    & 2                                           & 47                                              & ~~50.617                                         & 15                                                    \\ \hline
6/60                                           & 2                                       & 51                                          & 2                      & 58                         & ~~50.601                    & 2                                           & 47                                              & ~~50.600                                         & 30                          \\ \hline      
\hline                   
\end{tabular}
\caption{Addressing \qsix: approximate tiling problem  solution using specialized native algorithm for tile candidate generation and ASP encoding for selecting the resulting tiling (Divorce dataset)}
\label{til:div}
\end{table}

\begin{table}[]
\centering
\begin{tabular}{rrrrrrrrrr}
\hline
\hline
\multicolumn{1}{c}{\multirow{3}{*}{$n$/$\sigma$}} & \multicolumn{2}{c}{\revision{greedy}}                                                           & \multicolumn{7}{c}{hybrid}                                                                                                                                                                                                                                                              \\ \cline{2-10} 
\multicolumn{1}{c}{}                         & \multicolumn{1}{c}{\multirow{2}{*}{$k$}} & \multicolumn{1}{c}{\multirow{2}{*}{error}} & \multicolumn{3}{c}{first}                                                      & \multicolumn{1}{c}{\multirow{2}{*}{$k$ opt}} & \multicolumn{1}{c}{\multirow{2}{*}{error opt}} & \multicolumn{1}{c}{\multirow{2}{*}{time all}} & \multicolumn{1}{c}{\multirow{2}{*}{$|$all solutions$|$}} \\ \cline{4-6}
\multicolumn{1}{c}{}                         & \multicolumn{1}{c}{}                   & \multicolumn{1}{c}{}                       & \multicolumn{1}{c}{$k$} & \multicolumn{1}{c}{error} & \multicolumn{1}{c}{time} & \multicolumn{1}{c}{}                       & \multicolumn{1}{c}{}                           & \multicolumn{1}{c}{}                          & \multicolumn{1}{c}{}                                 \\ \hline
3/214                                          & 1                                       & 184                                         & 1                      & 196                        & ~~294.172                   & 3                                           & 88                                              & ~~294.173                                        & 7                                                     \\ \hline
4/168                                          & 3                                       & 142                                         & 2                      & 154                        & ~~1194.088                  & 3                                           & 142                                             & ~~1194.093                                       & 2                                                     \\ \hline
5/222                                          & 3                                       & 180                                         & 3                      & 192                        & ~~2126.944                  & 4                                           & 142                                             & ~~2126.948                                       & 11                                                    \\ \hline
6/224                                          & 3                                       & 182                                         & 3                      & 194                        & ~~2198.739                  & 5                                           & 142                                             & ~~2198.743                                       & 25  \\ \hline \hline                                                
\end{tabular}
\caption{Addressing \qsix: approximate tiling problem solution using specialized native algorithm for tile candidate generation and ASP encoding for selecting the resulting tiling (Glass dataset)}
\label{til:glass}
\end{table}


\section{Related Work}\label{sec:relwork}

Pattern mining approaches, especially frequent (closed or maximal) itemset, sequence, and subgraph mining are amongst the foundational methods in data mining (see, e.g., \citeN{aggarwal15data} for a recent textbook on the topic), and have been studied actively since mid-nineties \cite{agrawal93mining,DBLP:books/mit/fayyadPSU96/AgrawalMSTV96}. These problems are considered local and exhaustive, as the goal is always to enumerate all patterns that satisfy the (local) frequency constraint (together with some global constraints, such as closedness). In addition, the patterns have to be exact, that is, they have to be present in the data. The exactness was relaxed in later work \cite{pensa05towards}, while \citeN{tiling} considered the problem of summarizing the data using closed itemsests, that is, tiling. These two approaches, non-exact patterns and summarization using them, were combined by \citeN{miettinen08discrete} in their work on Boolean matrix factorization. 

The problem of enhancing pattern mining by injecting various user-specified constraints has recently gained increasing attention. On the one hand, optimized dedicated approaches exist, in which some of the constraints are deeply integrated into the mining algorithm (e.g., \citeNP{DBLP:conf/kdd/PeiH00}).  On the other hand, 
declarative methods based on Constraint Programming \cite{sky2014,DBLP:conf/cpaior/NegrevergneG15,DBLP:journals/corr/MetivierLC13}, SAT solving \cite{DBLP:conf/pakdd/JabbourSS15,DBLP:conf/cikm/JabbourSS13} and ASP \cite{DBLP:conf/lpnmr/Jarvisalo11,DBLP:conf/ijcai/GebserGQ0S16,DBLP:journals/corr/GuyetMQ14} have been proposed. 

Techniques from the last group are the closest to our work. However, in contrast to our method, they typically focus only on one particular pattern type and consider local constraints and condensed representations in isolation \cite{DBLP:conf/dmkd/PeiHM00,clospan}. 
\citeN{dp2013} and \citeN{DBLP:journals/ai/GunsDNTR17} focused on CP-based rather than ASP-based itemset mining and did not take into account sequences unlike we do. \citeN{DBLP:conf/ijcai/GebserGQ0S16} studied declarative sequence mining with ASP, but in contrast to our approach, optimized algorithms for frequent pattern discovery are not exploited in their method.
A theoretical framework for structured pattern mining was proposed by \citeN{DBLP:conf/aaai/GunsPN16}, whose main goal was to formally define the core components of the main mining tasks and compare dedicated mining algorithms to their declarative versions. While generic, this work did not take into account local and global constraints and neither has it been implemented.

\citeN{DBLP:conf/lpnmr/Jarvisalo11} and \citeN{DBLP:conf/ijcai/GebserGQ0S16} considered purely declarative ASP methods; unlike our approach, they do not admit integration of optimized mining algorithms and thus 
lack 
practicality. In fact, the need for such an integration in the context of complex structured mining was even explicitly stated
by \citeN{query_mining_ilp} and by \citeN{KR_Graphs}, which study formalizations of graph mining problems using logical means. \revision{While the ASP \cite{DBLP:conf/ijcai/GebserGQ0S16} and CP models \cite{DBLP:conf/cpaior/NegrevergneG15} for frequent pattern mining cannot be hybridized completely out-of-the-box, due to dependencies between constraints and assumptions on the input-output structure, in principle, as we see from our work that it is possible to re-use ideas and the general modelling approach to turn them into hybrid models to reach more practical performance, for example, our ASP model for sequence mining is inspired by the non-hybrid ASP model of \citeN{DBLP:conf/ijcai/GebserGQ0S16}. The main observation here is that typically these systems have a number of constraint groups: to generate patterns, check patterns validity and perform dominance or group-property checks. Our hybridization idea suggests to replace one of these groups with a highly optimized solver, while keeping the other groups (with minor changes to adapt for the input-output structure) in the model to guarantee generality.}

\section{Conclusion}\label{sec:conc}

We have presented a hybrid approach for condensed itemset, sequence and graph mining, which uses the optimized dedicated algorithms to determine the frequent patterns and post-filters them using a declarative ASP program. The idea of exploiting ASP for pattern mining is not new; it was studied for both itemsets and sequences. However, unlike previous methods we made steps towards optimizing the declarative techniques by making use of the existing specialized methods and also integrated the dominance programming machinery in our implementation to allow for combining local and global constraints on a generic level. Moreover, using the example of the approximate tile selection problem, we have demonstrated that our hybrid method can be further generalized to other data mining tasks.

One of the possible future directions is to extend the proposed approach to an iterative technique, where dedicated data mining and declarative methods are interlinked and applied in an alternating fashion. More specifically, all constraints can be split into two parts: those that can be effectively handled using declarative means and those for which specialized algorithms are much more scalable. Answer set programs with external computations \cite{hp} could be possibly exploited in this mining context. 

Another promising but challenging research stream concerns the integration of data \emph{decomposition} techniques into our approach. Here, one can divide a given dataset into several parts, such that the frequent patterns are identified in these parts separately, and then the results are effectively combined; such data decomposition is expected to yield further computational gains. 

\paragraph{Acknowledgment.} This work has been supported by the
FWO and by the ERC-ADG-201 project 694980 SYNTH
funded by the European Research Council.


\appendix
\section{Correctness of ASP Encoding for Maximal Itemset Computation}
\noindent\textbf{Proposition.} Let $S$ be a set of frequent itemsets, represented as facts of the form \texttt{item(i,j)} reflecting that the itemset with ID \texttt{i} contains an item \texttt{j}.
Moreover, let \texttt{valid(i)} be in $S$ if \texttt{i} is a valid pattern, and let $P$ be the following logic program:
\medskip

\noindent\small{\texttt{(1)\; 1 \{selected(I) : valid(I)\} 1.\\
(2)\;not\_superset(J) :- selected(I), item(I,V), not item(J,V), valid(J), I != J.\\
(3)\;dominated :- selected(I), valid(J), I != J, not not\_superset(J).\\
(4)\;:- dominated.
}
}
\medskip

Then it holds that 
\begin{itemize}
\item[(i)] if $I\supseteq $\texttt{selected(x)} is an answer set of $P\cup S$, then $x$ is a valid maximal itemset 
\item[(ii)] if an itemset with ID \texttt{y} is a valid maximal itemset then an answer set $I$ of $P\cup S$ exists, such that \texttt{selected(y)}$\in I$.
\end{itemize}
\bigskip

\noindent \textbf{Proof.} \begin{itemize}
\item[(i)] Let $I\supseteq $\texttt{selected(x)} be an answer set of $P\cup S$. Since $S$ does not contain any facts over the \texttt{selected} predicate by our assumption, we have that \texttt{selected(x)} must have been obtained due to the rule (1), i.e., \texttt{valid(x)}$\in S$, meaning that the itemset \texttt{x} is valid. Since the set $S$ contains only frequent itemsets, \texttt{x} must be also frequent. Moreover, we know that $S$ also stores all other frequent itemsets. Therefore, in order to show that \texttt{x} is maximal, we need to show that no valid itemset in $S$ is a superset of \texttt{x}. Towards a contradiction, assume the contrary, and let \texttt{x'} be such an itemset. Then it holds that every item that is in \texttt{x} is also contained in \texttt{x'}. However, in this case the body of the grounding of the rule (2) with \texttt{I} substituted by \texttt{x} and \texttt{J} by \texttt{x'} is not satisfied. Thus, \texttt{not\_superset(x')}$\not \in I$, as the respective predicate does not appear in any other rule head of $P$. However, then the body of (3) is satisfied for the grounding with \texttt{I} and \texttt{J} being substituted by \texttt{x} and \texttt{x'} respectively, meaning that \texttt{dominated} must be present in $I$, for it to be a model of $P \cup S$, but then the constraint (4) is violated, leading to the contradiction of $I$ being a model of $P\cup S$.
\item[(ii)] Suppose that \texttt{x} is a valid maximal itemset. Then by construction of $S$, \texttt{valid(x)}$\in S$. Consider an interpretation $I=S \cup \{\texttt{selected(x),valid(x)}\cup\{\texttt{not\_superset(x')\,|\,valid(x')}\in S, \texttt{x'}\neq \texttt{x}\}\}$. We show that this interpretation is an answer set of $P\cup S$. Towards a contradiction, assume the contrary. Then either (a) $I$ is not a model of $P \cup S$ or (b) $I$ is not minimal. First suppose that (a) holds. Then we have that $body(r) \in I$, but $\mi{head(r)\not \in I}$ for some rule $r$ among the rules (1)-(4). Note that since by construction, $I$ contains just a single atom over the \texttt{selected} predicate, i.e., \texttt{selected(x)}, such that \texttt{valid(x)}$\in S$, the rule (1) must be satisfied. Assume that (2) is not satisfied. Then the itemset \texttt{x} must contain some item \texttt{j}, which another valid itemset \texttt{x'} does not contain, but \texttt{not\_superset(x')}$\not \in I$. By construction of $I$, we have \texttt{not\_superset(x)} for all valid itemsets in $S$ apart from \texttt{x}; in particular, \texttt{not\_superset(x')}$\in I$, meaning that (2) is satisfied. Therefore, (3) must be a problematic rule. However, since by construction \texttt{not\_superset(x')}$\in I$ for all \texttt{x'} such that \texttt{valid(x')}$\in S$, the body of (3) cannot be satisfied. Finally, again by construction \texttt{dominated}$\not \in I$. This means that $I$ is a model of $P \cup S$. Hence, (b) must hold, i.e., $I'\subseteq I$ must exists, such that $I'$ is a model of $P \cup S$. Note that for $I'$ to be a model, it must contain all facts in $S$. The assumption that $\texttt{selected(x)}\in I\backslash I'$ is not valid by construction of $I$. If $\texttt{not\_superset(x')} \in I \backslash I'$ for some \texttt{x'}, such that \texttt{valid(x')}$\in S$ then the body of rule (3) is satisfied by $I'$ but not its head, as \texttt{dominated}$\not \in I$ by construction. This means that $I$ must be a minimal model of $P \cup S$ and thus its answer set. \qed
\end{itemize}

\bibliographystyle{style/acmtrans}
\bibliography{biblio}

\begin{thebibliography}{}

\bibitem[\protect\citeauthoryear{Aggarwal}{Aggarwal}{2015}]{aggarwal15data}
{\sc Aggarwal, C.~C.} 2015.
\newblock {\em Data Mining: The Textbook}.
\newblock Springer, Cham.

\bibitem[\protect\citeauthoryear{Agrawal, Imielinski, and Swami}{Agrawal
  et~al\mbox{.}}{1993}]{agrawal93mining}
{\sc Agrawal, R.}, {\sc Imielinski, T.}, {\sc and} {\sc Swami, A.} 1993.
\newblock Mining association rules between sets of items in large databases.
\newblock In {\em SIGMOD '93}. 207--216.

\bibitem[\protect\citeauthoryear{Agrawal, Mannila, Srikant, Toivonen, and
  Verkamo}{Agrawal
  et~al\mbox{.}}{1996}]{DBLP:books/mit/fayyadPSU96/AgrawalMSTV96}
{\sc Agrawal, R.}, {\sc Mannila, H.}, {\sc Srikant, R.}, {\sc Toivonen, H.},
  {\sc and} {\sc Verkamo, A.~I.} 1996.
\newblock Fast discovery of association rules.
\newblock In {\em Advances in Knowledge Discovery and Data Mining}. {AAAI/MIT}
  Press, 307--328.

\bibitem[\protect\citeauthoryear{Alviano, Dodaro, Faber, Leone, and
  Ricca}{Alviano et~al\mbox{.}}{2013}]{wasp}
{\sc Alviano, M.}, {\sc Dodaro, C.}, {\sc Faber, W.}, {\sc Leone, N.}, {\sc
  and} {\sc Ricca, F.} 2013.
\newblock Wasp: A native asp solver based on constraint learning.
\newblock In {\em International Conference on Logic Programming and
  Nonmonotonic Reasoning}. Springer, 54--66.

\bibitem[\protect\citeauthoryear{Aoga, Guns, and Schaus}{Aoga
  et~al\mbox{.}}{2016}]{PPIC}
{\sc Aoga, J. O.~R.}, {\sc Guns, T.}, {\sc and} {\sc Schaus, P.} 2016.
\newblock An efficient algorithm for mining frequent sequence with constraint
  programming.
\newblock In {\em {ECML} {PKDD}}. 315--330.

\bibitem[\protect\citeauthoryear{Babai}{Babai}{2015}]{graph_isomorphism}
{\sc Babai, L.} 2015.
\newblock Graph isomorphism in quasipolynomial time.
\newblock {\em CoRR\/}~{\em abs/1512.03547}.

\bibitem[\protect\citeauthoryear{Bonchi and Lucchese}{Bonchi and
  Lucchese}{2006}]{DBLP:journals/kais/BonchiL06}
{\sc Bonchi, F.} {\sc and} {\sc Lucchese, C.} 2006.
\newblock On condensed representations of constrained frequent patterns.
\newblock {\em Knowl. Inf. Syst.\/}~{\em 9,\/}~2, 180--201.

\bibitem[\protect\citeauthoryear{Cook}{Cook}{1971}]{DBLP:conf/stoc/Cook71}
{\sc Cook, S.~A.} 1971.
\newblock The complexity of theorem-proving procedures.
\newblock In {\em Proceedings of the 3rd Annual {ACM} Symposium on Theory of
  Computing, May 3-5, 1971, Shaker Heights, Ohio, {USA}}, {M.~A. Harrison},
  {R.~B. Banerji}, {and} {J.~D. Ullman}, Eds. {ACM}, 151--158.

\bibitem[\protect\citeauthoryear{Eiter, Brewka, Dao{-}Tran, Fink, Ianni, and
  Krennwallner}{Eiter et~al\mbox{.}}{2009}]{hp}
{\sc Eiter, T.}, {\sc Brewka, G.}, {\sc Dao{-}Tran, M.}, {\sc Fink, M.}, {\sc
  Ianni, G.}, {\sc and} {\sc Krennwallner, T.} 2009.
\newblock Combining nonmonotonic knowledge bases with external sources.
\newblock In {\em Frontiers of Combining Systems, 7th International Symposium,
  FroCoS Italy, September 16-18}. 18--42.

\bibitem[\protect\citeauthoryear{Eiter, Ianni, and Krennwallner}{Eiter
  et~al\mbox{.}}{2009}]{eiter}
{\sc Eiter, T.}, {\sc Ianni, G.}, {\sc and} {\sc Krennwallner, T.} 2009.
\newblock Answer set programming: A primer.
\newblock In {\em 5th International Reasoning Web Summer School (RW 2009),
  Brixen/Bressanone, Italy, August 30 -- September 4, 2009}. LNCS, vol. 5689.
  Springer.

\bibitem[\protect\citeauthoryear{Faber, Pfeifer, Leone, Dell'Armi, and
  Ielpa}{Faber et~al\mbox{.}}{2008}]{dlv_aggregates}
{\sc Faber, W.}, {\sc Pfeifer, G.}, {\sc Leone, N.}, {\sc Dell'Armi, T.}, {\sc
  and} {\sc Ielpa, G.} 2008.
\newblock Design and implementation of aggregate functions in the {DLV} system.
\newblock {\em CoRR\/}~{\em abs/0802.3137}.

\bibitem[\protect\citeauthoryear{Gebser, Guyet, Quiniou, Romero, and
  Schaub}{Gebser et~al\mbox{.}}{2016}]{DBLP:conf/ijcai/GebserGQ0S16}
{\sc Gebser, M.}, {\sc Guyet, T.}, {\sc Quiniou, R.}, {\sc Romero, J.}, {\sc
  and} {\sc Schaub, T.} 2016.
\newblock Knowledge-based sequence mining with {ASP}.
\newblock In {\em Proceedings of the Twenty-Fifth International Joint
  Conference on Artificial Intelligence, {IJCAI} 2016, New York, NY, USA, 9-15
  July 2016}.

\bibitem[\protect\citeauthoryear{Gebser, Kaminski, Kaufmann, and Schaub}{Gebser
  et~al\mbox{.}}{2012}]{ASPbook}
{\sc Gebser, M.}, {\sc Kaminski, R.}, {\sc Kaufmann, B.}, {\sc and} {\sc
  Schaub, T.} 2012.
\newblock {\em Answer Set Solving in Practice}.
\newblock Synthesis Lectures on Artificial Intelligence and Machine Learning.
  Morgan and Claypool Publishers.

\bibitem[\protect\citeauthoryear{Gebser, Kaufmann, Neumann, and Schaub}{Gebser
  et~al\mbox{.}}{2007}]{clasp}
{\sc Gebser, M.}, {\sc Kaufmann, B.}, {\sc Neumann, A.}, {\sc and} {\sc Schaub,
  T.} 2007.
\newblock {\it clasp} : A conflict-driven answer set solver.
\newblock In {\em LPNMR}. 260--265.

\bibitem[\protect\citeauthoryear{Geerts, Goethals, and
  Mielik{\"{a}}inen}{Geerts et~al\mbox{.}}{2004}]{tiling}
{\sc Geerts, F.}, {\sc Goethals, B.}, {\sc and} {\sc Mielik{\"{a}}inen, T.}
  2004.
\newblock Tiling databases.
\newblock In {\em Discovery Science, 7th International Conference, {DS} 2004,
  Padova, Italy, October 2-5, 2004, Proceedings}. 278--289.

\bibitem[\protect\citeauthoryear{Gelfond and Lifschitz}{Gelfond and
  Lifschitz}{1988}]{GL1988}
{\sc Gelfond, M.} {\sc and} {\sc Lifschitz, V.} 1988.
\newblock The stable model semantics for logic programming.
\newblock In {\em Proc. of ICLP/SLP}. 1070--1080.

\bibitem[\protect\citeauthoryear{Guns, Dries, Nijssen, Tack, and De~Raedt}{Guns
  et~al\mbox{.}}{2017}]{DBLP:journals/ai/GunsDNTR17}
{\sc Guns, T.}, {\sc Dries, A.}, {\sc Nijssen, S.}, {\sc Tack, G.}, {\sc and}
  {\sc De~Raedt, L.} 2017.
\newblock {MiningZinc}: {A} declarative framework for constraint-based mining.
\newblock {\em Artif. Intell.\/}~{\em 244}, 6--29.

\bibitem[\protect\citeauthoryear{Guns, Nijssen, and {de Raedt}}{Guns
  et~al\mbox{.}}{2013}]{k_pattern_mining_under_constraints}
{\sc Guns, T.}, {\sc Nijssen, S.}, {\sc and} {\sc {de Raedt}, L.} 2013.
\newblock k-pattern set mining under constraints.
\newblock {\em IEEE Trans. on Knowl. and Data Eng.\/}~{\em 25,\/}~2 (Feb.),
  402--418.

\bibitem[\protect\citeauthoryear{Guns, Paramonov, and N{\'{e}}grevergne}{Guns
  et~al\mbox{.}}{2016}]{DBLP:conf/aaai/GunsPN16}
{\sc Guns, T.}, {\sc Paramonov, S.}, {\sc and} {\sc N{\'{e}}grevergne, B.}
  2016.
\newblock On declarative modeling of structured pattern mining.
\newblock In {\em Declarative Learning Based Programming, Papers from the 2016
  {AAAI} Workshop, Phoenix, Arizona, USA, February 13, 2016.}

\bibitem[\protect\citeauthoryear{Guyet, Moinard, and Quiniou}{Guyet
  et~al\mbox{.}}{2014}]{DBLP:journals/corr/GuyetMQ14}
{\sc Guyet, T.}, {\sc Moinard, Y.}, {\sc and} {\sc Quiniou, R.} 2014.
\newblock Using answer set programming for pattern mining.
\newblock {\em CoRR\/}~{\em abs/1409.7777}.

\bibitem[\protect\citeauthoryear{Jabbour, Sais, and Salhi}{Jabbour
  et~al\mbox{.}}{2013}]{DBLP:conf/cikm/JabbourSS13}
{\sc Jabbour, S.}, {\sc Sais, L.}, {\sc and} {\sc Salhi, Y.} 2013.
\newblock Boolean satisfiability for sequence mining.
\newblock In {\em 22nd {ACM} International Conference on Information and
  Knowledge Management, CIKM'13, San Francisco, CA, USA, October 27 - November
  1, 2013}. 649--658.

\bibitem[\protect\citeauthoryear{Jabbour, Sais, and Salhi}{Jabbour
  et~al\mbox{.}}{2015}]{DBLP:conf/pakdd/JabbourSS15}
{\sc Jabbour, S.}, {\sc Sais, L.}, {\sc and} {\sc Salhi, Y.} 2015.
\newblock Decomposition based {SAT} encodings for itemset mining problems.
\newblock In {\em {PAKDD}}. 662--674.

\bibitem[\protect\citeauthoryear{J{\"{a}}rvisalo}{J{\"{a}}rvisalo}{2011}]{DBLP:conf/lpnmr/Jarvisalo11}
{\sc J{\"{a}}rvisalo, M.} 2011.
\newblock Itemset mining as a challenge application for answer set enumeration.
\newblock In {\em Logic Programming and Nonmonotonic Reasoning - 11th
  International Conference, {LPNMR} 2011, Vancouver, Canada, May 16-19, 2011.
  Proceedings}. 304--310.

\bibitem[\protect\citeauthoryear{Kimelfeld and Kolaitis}{Kimelfeld and
  Kolaitis}{2014}]{kimelfeld14complexity}
{\sc Kimelfeld, B.} {\sc and} {\sc Kolaitis, P.~G.} 2014.
\newblock {The Complexity of Mining Maximal Frequent Subgraphs}.
\newblock {\em ACM Trans. Database Syst.\/}~{\em 39,\/}~4, 32--33.

\bibitem[\protect\citeauthoryear{Lifschitz}{Lifschitz}{2008}]{whatisasp}
{\sc Lifschitz, V.} 2008.
\newblock What is answer set programming?
\newblock AAAI.

\bibitem[\protect\citeauthoryear{Mannila and Toivonen}{Mannila and
  Toivonen}{1997}]{DBLP:journals/datamine/MannilaT97}
{\sc Mannila, H.} {\sc and} {\sc Toivonen, H.} 1997.
\newblock Levelwise search and borders of theories in knowledge discovery.
\newblock {\em Data Min. Knowl. Discov.\/}~{\em 1,\/}~3, 241--258.

\bibitem[\protect\citeauthoryear{M{\'{e}}tivier, Loudni, and
  Charnois}{M{\'{e}}tivier
  et~al\mbox{.}}{2013}]{DBLP:journals/corr/MetivierLC13}
{\sc M{\'{e}}tivier, J.}, {\sc Loudni, S.}, {\sc and} {\sc Charnois, T.} 2013.
\newblock A constraint programming approach for mining sequential patterns in a
  sequence database.
\newblock {\em CoRR\/}~{\em abs/1311.6907}.

\bibitem[\protect\citeauthoryear{Miettinen}{Miettinen}{2008}]{miettinen08positive-negative}
{\sc Miettinen, P.} 2008.
\newblock On the positive-negative partial set cover problem.
\newblock {\em Inform. Process. Lett.\/}~{\em 108,\/}~4, 219--221.

\bibitem[\protect\citeauthoryear{Miettinen}{Miettinen}{2009}]{miettinen09matrix}
{\sc Miettinen, P.} 2009.
\newblock Matrix decomposition methods for data mining: {Computational}
  complexity and algorithms.
\newblock Ph.D. thesis, Department of Computer Science, University of Helsinki.

\bibitem[\protect\citeauthoryear{Miettinen}{Miettinen}{2015}]{miettinen15generalized}
{\sc Miettinen, P.} 2015.
\newblock Generalized matrix factorizations as a unifying framework for pattern
  set mining: {Complexity} beyond blocks.
\newblock In {\em ECMLPKDD '15}. 36--52.

\bibitem[\protect\citeauthoryear{Miettinen, Mielik{\"a}inen, Gionis, Das, and
  Mannila}{Miettinen et~al\mbox{.}}{2008}]{miettinen08discrete}
{\sc Miettinen, P.}, {\sc Mielik{\"a}inen, T.}, {\sc Gionis, A.}, {\sc Das,
  G.}, {\sc and} {\sc Mannila, H.} 2008.
\newblock The discrete basis problem.
\newblock {\em IEEE Trans. Knowl. Data Eng.\/}~{\em 20,\/}~10, 1348--1362.

\bibitem[\protect\citeauthoryear{N{\'{e}}grevergne, Dries, Guns, and
  Nijssen}{N{\'{e}}grevergne et~al\mbox{.}}{2013}]{dp2013}
{\sc N{\'{e}}grevergne, B.}, {\sc Dries, A.}, {\sc Guns, T.}, {\sc and} {\sc
  Nijssen, S.} 2013.
\newblock Dominance programming for itemset mining.
\newblock In {\em 2013 {IEEE} 13th International Conference on Data Mining,
  Dallas, TX, USA, December 7-10, 2013}. 557--566.

\bibitem[\protect\citeauthoryear{N{\'{e}}grevergne and Guns}{N{\'{e}}grevergne
  and Guns}{2015}]{DBLP:conf/cpaior/NegrevergneG15}
{\sc N{\'{e}}grevergne, B.} {\sc and} {\sc Guns, T.} 2015.
\newblock Constraint-based sequence mining using constraint programming.
\newblock In {\em {CPAIOR}}. 288--305.

\bibitem[\protect\citeauthoryear{Neumann and Miettinen}{Neumann and
  Miettinen}{2017}]{neumann17reductions}
{\sc Neumann, S.} {\sc and} {\sc Miettinen, P.} 2017.
\newblock {Reductions for Frequency-Based Data Mining Problems}.
\newblock In {\em Proceedings of the 2017 IEEE International Conference on Data
  Mining}. 997--1002.

\bibitem[\protect\citeauthoryear{Paramonov, van Leeuwen, Denecker, and
  De~Raedt}{Paramonov et~al\mbox{.}}{2015}]{query_mining_ilp}
{\sc Paramonov, S.}, {\sc van Leeuwen, M.}, {\sc Denecker, M.}, {\sc and} {\sc
  De~Raedt, L.} 2015.
\newblock An exercise in declarative modeling for relational query mining.
\newblock In {\em {ILP}}. 166--182.

\bibitem[\protect\citeauthoryear{Pei and Han}{Pei and
  Han}{2000}]{DBLP:conf/kdd/PeiH00}
{\sc Pei, J.} {\sc and} {\sc Han, J.} 2000.
\newblock Can we push more constraints into frequent pattern mining?
\newblock In {\em {ACM} {SIGKDD}, Boston, MA, USA}. 350--354.

\bibitem[\protect\citeauthoryear{Pei, Han, and Mao}{Pei
  et~al\mbox{.}}{2000}]{DBLP:conf/dmkd/PeiHM00}
{\sc Pei, J.}, {\sc Han, J.}, {\sc and} {\sc Mao, R.} 2000.
\newblock {CLOSET:} an efficient algorithm for mining frequent closed itemsets.
\newblock In {\em {ACM} {SIGMOD} Workshop on Research Issues in Data Mining and
  Knowledge Discovery, Dallas, TX, USA}. 21--30.

\bibitem[\protect\citeauthoryear{Pensa and Boulicaut}{Pensa and
  Boulicaut}{2005}]{pensa05towards}
{\sc Pensa, R.~G.} {\sc and} {\sc Boulicaut, J.-F.} 2005.
\newblock Towards fault-tolerant formal concept analysis.
\newblock In {\em AI*IA '05}. 212--223.

\bibitem[\protect\citeauthoryear{Rojas, Boizumault, Loudni, Cr{\'{e}}milleux,
  and Lepailleur}{Rojas et~al\mbox{.}}{2014}]{sky2014}
{\sc Rojas, W.~U.}, {\sc Boizumault, P.}, {\sc Loudni, S.}, {\sc
  Cr{\'{e}}milleux, B.}, {\sc and} {\sc Lepailleur, A.} 2014.
\newblock Mining (soft-) skypatterns using dynamic {CSP}.
\newblock In {\em {CPAIOR}}. 71--87.

\bibitem[\protect\citeauthoryear{Simons, Niemel{\"{a}}, and Soininen}{Simons
  et~al\mbox{.}}{2002}]{DBLP:journals/ai/SimonsNS02}
{\sc Simons, P.}, {\sc Niemel{\"{a}}, I.}, {\sc and} {\sc Soininen, T.} 2002.
\newblock Extending and implementing the stable model semantics.
\newblock {\em Artif. Intell.\/}~{\em 138,\/}~1-2, 181--234.

\bibitem[\protect\citeauthoryear{Tyukin, Kramer, and Wicker}{Tyukin
  et~al\mbox{.}}{2014}]{tyukin14bmad}
{\sc Tyukin, A.}, {\sc Kramer, S.}, {\sc and} {\sc Wicker, J.} 2014.
\newblock {BMaD} {\textendash} {A} boolean matrix decomposition framework.
\newblock In {\em ECML PKDD '14}, {T.~Calders}, {F.~Esposito},
  {E.~H{\"u}llermeier}, {and} {R.~Meo}, Eds. 481--484.

\bibitem[\protect\citeauthoryear{van~der Hallen, Paramonov, Leuschel, and
  Janssens}{van~der Hallen et~al\mbox{.}}{2016}]{KR_Graphs}
{\sc van~der Hallen, M.}, {\sc Paramonov, S.}, {\sc Leuschel, M.}, {\sc and}
  {\sc Janssens, G.} 2016.
\newblock Knowledge representation analysis of graph mining.
\newblock {\em CoRR\/}~{\em abs/1608.08956}.

\bibitem[\protect\citeauthoryear{Yan and Han}{Yan and Han}{2002}]{gspan}
{\sc Yan, X.} {\sc and} {\sc Han, J.} 2002.
\newblock gspan: Graph-based substructure pattern mining.
\newblock In {\em ICDM '02}.

\bibitem[\protect\citeauthoryear{Yan, Han, and Afshar}{Yan
  et~al\mbox{.}}{2003}]{clospan}
{\sc Yan, X.}, {\sc Han, J.}, {\sc and} {\sc Afshar, R.} 2003.
\newblock Clospan: Mining closed sequential patterns in large datasets.
\newblock In {\em In SDM}. 166--177.

\bibitem[\protect\citeauthoryear{Zaki, Parthasarathy, Ogihara, and Li}{Zaki
  et~al\mbox{.}}{1997}]{eclat}
{\sc Zaki, M.~J.}, {\sc Parthasarathy, S.}, {\sc Ogihara, M.}, {\sc and} {\sc
  Li, W.} 1997.
\newblock New algorithms for fast discovery of association rules.
\newblock Tech. rep., Rochester, NY, USA.

\end{thebibliography}

\label{lastpage}
\end{document}